\documentclass[11pt]{article}

\usepackage[margin=1in]{geometry}
\usepackage[utf8]{inputenc}
\usepackage[T1]{fontenc}
\usepackage{lmodern}
\usepackage[dvipsnames]{xcolor}     
\usepackage{pifont}   
\usepackage{multirow}
\usepackage{booktabs} 
\usepackage{siunitx}  
\usepackage{fancyvrb}
\usepackage{xcolor}
\definecolor{tiiPurple}{RGB}{122, 0, 255}

\usepackage{hyperref}
\usepackage{subcaption} 
\hypersetup{
    colorlinks=true,
    linkcolor=tiiPurple,
    urlcolor=tiiPurple,
    citecolor=tiiPurple
}

\definecolor{GoodGain}{HTML}{2ca02c} 
\definecolor{BadGain}{HTML}{d62728}  

\usepackage[most]{tcolorbox}
\usepackage{graphicx}
\usepackage{tabularx}
\usepackage[useregional]{datetime2}
\DTMsetdatestyle{iso}  
\usepackage[normalem]{ulem}
\usepackage{titlesec}
\usepackage{adjustbox}
\usepackage{enumitem}
\usepackage{booktabs}
\usepackage{multirow}
\usepackage{array}
\usepackage{float}
\usepackage{caption}
\usepackage{amsfonts}       
\usepackage{amsmath, amssymb}       
\usepackage{nicefrac}       
\usepackage{hyperref}       
\usepackage{url}            
\usepackage[utf8]{inputenc} 
\usepackage[T1]{fontenc}    
\usepackage{natbib}
\usepackage{graphicx}
\usepackage{enumitem}
\usepackage{makecell}
\definecolor{bestcolor}{RGB}{220,255,220}
\usepackage{CJKutf8}
\usepackage[para]{threeparttable}

\titleformat{\section}
  {\normalfont\Large\bfseries}{\thesection.}{1em}{}
\usepackage{fancyhdr}
\renewcommand{\headrulewidth}{0.4pt}  
\begin{document}
\noindent
\begin{minipage}[t]{0.49\textwidth}
    \vspace*{-4.2em}
    \includegraphics[height=1.3cm,width=2.5cm]{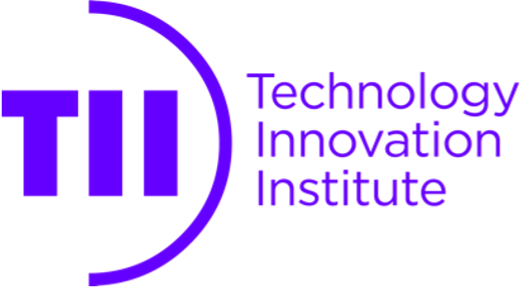}  
\end{minipage}%
\hfill
\begin{minipage}[t]{0.49\textwidth}
    \vspace*{-1.5em}
    \raggedleft
    \today
\end{minipage}
\vspace*{-0.5em}
\hrule
\vspace{1.2em}

\begin{center}
    {\Large \textbf{\textcolor{tiiPurple}{Learnable Multipliers: Freeing the Scale of Language Model Matrix Layers}}}
\end{center}
\noindent

{%
  \renewcommand\thefootnote{\fnsymbol{footnote}}%
  \footnotetext[1]{Equal contribution.}%
}

\begin{center}

Maksim Velikanov\textsuperscript{*}, Ilyas Chahed\textsuperscript{*}, Jingwei Zuo, Dhia Eddine Rhaiem, \\
Younes Belkada, Hakim Hacid \\
\newcommand{\equalcontrib}{\thanks{Equal contribution.}}
\vspace{1em}
\textbf{Falcon LLM Team} \\

\begin{tabular}{@{}l l@{}}
    \raisebox{-0.15\height}{\includegraphics[width=0.4cm]{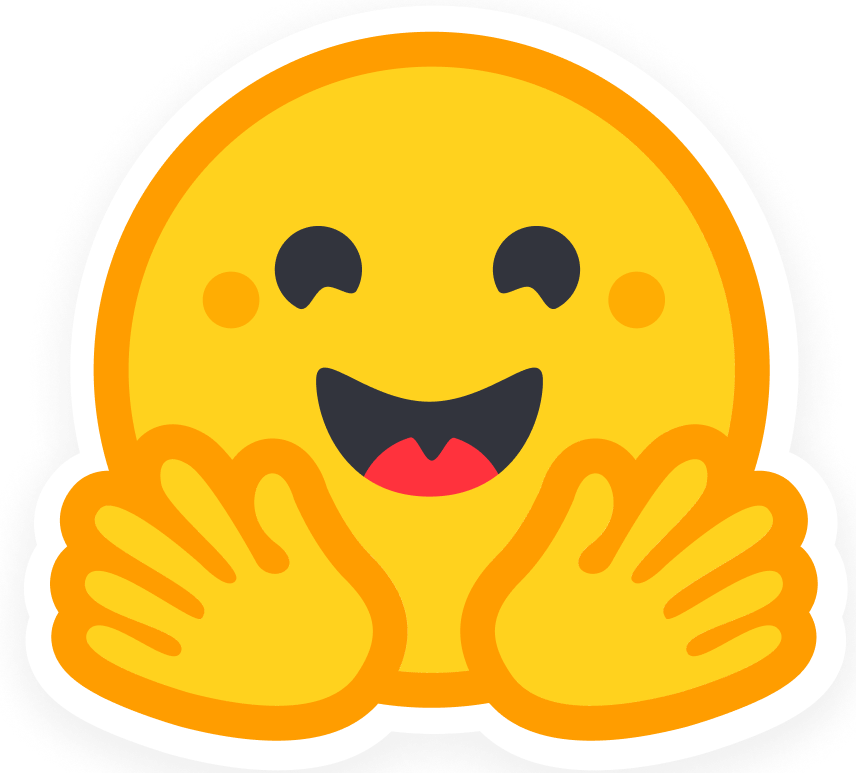}} &
    \href{https://huggingface.co/tiiuae}{https://huggingface.co/tiiuae} 
\end{tabular}

\end{center}


\begin{abstract}
Applying weight decay (WD) to matrix layers is standard practice in large-language-model pretraining. Prior work suggests that stochastic gradient noise induces a Brownian-like expansion of the weight matrices $W$, whose growth is counteracted by WD, leading to a WD-noise equilibrium with a certain weight norm $\|W\|$. In this work, we view the equilibrium norm as a harmful artifact of the training procedure, and address it by introducing \textit{learnable multipliers} to learn the optimal scale. First, we attach a learnable scalar multiplier to $W$ and confirm that the WD–noise equilibrium norm is suboptimal: the learned scale adapts to data and improves performance. We then argue that individual row and column norms are similarly constrained, and free their scale by introducing learnable per-row and per-column multipliers. 
Our method can be viewed as a learnable, more expressive generalization of $\mu$P multipliers. It outperforms a well-tuned $\mu$P baseline, reduces the computational overhead of multiplier tuning, and surfaces practical questions such as forward-pass symmetries and the width-scaling of the learned multipliers.
Finally, we validate learnable multipliers with both Adam and Muon optimizers, where it shows improvement in downstream evaluations matching the improvement of the switching from Adam to Muon.     
\end{abstract}

\section{Introduction}
Pretraining large-scale language models presents significant challenges for both the optimization algorithm and the choice of hyperparameters. The most widely used and reliable optimizer is Adam \citep{Adam_KingmaB14}, or, rather, its weight decay version AdamW \citep{loshchilov2018decoupled} that incorporates decay directly in the parameter update: $\boldsymbol{\theta}_t \leftarrow \boldsymbol{\theta}_t-\eta\lambda\boldsymbol{\theta}_t$, where $\eta$ is the learning rate (LR) and $\lambda$ the weight decay hyperparameter. The WD term of AdamW is critical for both improving model performance and stabilizing training at large scale \citep{Devlin2019BERTPO,NEURIPS2020_1457c0d6}. Recent alternatives to Adam, such as Muon \citep{Keller2024muon}, similarly rely on the explicit WD term in its parameter update to maintain stability and performance \citep{liu2025muonscalablellmtraining}.  

The practical ubiquity and necessity of WD term motivates a deeper investigation into its effects on training dynamics that might reveal the reasons behind its success, or potential shortcomings. Prior work points to a wide range of WD effects on the training, from improving the bias-variance tradeoff \citep{Angelo2024whyweightdecay} to imposing certain structures in the model weights \citep{kobayashi2024weight}. Central to this work is the stochastic gradient noise perspective on weight decay \citep{kosson2024rotationalequilibriumweightdecay,zuo2025falconh1familyhybridheadlanguage}. From this viewpoint, the gradient noise induces a Brownian-like component to the optimizer updates that may cause uncontrollable growth of model weights if left unchecked. Weight decay counteracts this Brownian expansion, resulting in a \emph{noise-WD equilibrium}. In the equilibrium, the norm of the model weights $W$ scales predictably with the learning rate $\eta$ and WD $\lambda$, as can be both observed empirically and derived from toy models \citep{kosson2024rotationalequilibriumweightdecay,zuo2025falconh1familyhybridheadlanguage}:
\begin{equation}\label{eq:equilibrium_norm_scaling}
    \|W(\eta,\lambda)\|\propto S(\eta,\lambda)=\sqrt{\frac{\eta}{\lambda}}.
\end{equation}  
Our work builds on a straightforward interpretation of the equilibrium norm scaling \eqref{eq:equilibrium_norm_scaling}: the weight norms are dictated by the optimization hyperparameters instead of being learned from the data. In other words, WD traps model weights in the noise-driven equilibrium, preventing them from learning the scale suitable for a given training data. To escape the noise-WD equilibrium, we propose to re-parametrize the model weights with \emph{Learnable Multipliers (LRM)}, introduced in sec.~\ref{sec:learnable_multipliers}. For example, in the scalar case, $W\rightarrow sW$ with the learnable scale $s\in\mathbb{R}$. We expect learnable multipliers to not experience the same noise-WD problem by looking at the modern LLM practices: WD is not applied to scalar and vector-like weights, e.g. RMSNorm weights, without stability or performance issues caused by removing WD from matrix layers.
    
\paragraph{Overview of the results.} We first introduce scalar and vector learnable multipliers in section~\ref{sec:learnable_multipliers}. We provide a general discussion of their placement within language model architectures, their connection to maximal update parametrization ($\mu$P), and the instances of learnable multipliers in prior work. We then investigate LRMs in the context of language model pretraining.
\begin{itemize} 
    \item In section \ref{sec:whatlearned}, we confirm that features learned by matrix layers alone are limited by noise-WD equilibrium, while adding learnable multipliers results in richer representations. First, we design a series of experiments to show that the matrix layers fail to adapt their scale when it is required for optimal loss minimization, while adding multipliers recovers this lost performance. Then, we demonstrate that multipliers allow for a more diverse scale distribution across residual blocks. Likewise, vector LRMs enable a more diverse scale distribution for internal features within each model block.   
    \item In section \ref{sec:details}, we point to various aspects essential for stable and effective application of learnable multipliers to model pretraining. This includes handling the reparameterization symmetries of a given model architecture, exploring width $\mu$P scaling, and other details such as addressing interaction of multipliers with gradient clipping.  
    \item In section \ref{sec:results}, we validate the performance in a longer end-to-end pretraining run. Learnable multipliers maintain an increasing performance gap over the baseline throughout the whole training, supporting the earlier conclusion that multipliers enable richer model representation. 
    Finally, we explore the role of multiplier initialization using tuned values of $\mu$P multipliers from \citep{zuo2025falconh1familyhybridheadlanguage}. LRMs maintain the same level of performance regardless of whether the tuned values of forward and WD multipliers are used, while having tuned learning rate multipliers is still important for optimal performance.
    \item Learnable multipliers act on weight matrices in an isolated manner, suggesting a native application to a wide range of architectures and optimizers. We illustrate this applicability by doing experiments on a hybrid attention-SSM architecture and applying LRMs to structurally distinct attention, SSM, and MLP residual blocks. To illustrate optimizer applicability, we perform some of the experiments for Adam and Muon, showing similar performance gains and behavior patterns in both cases.
\end{itemize}

\paragraph{Notations.} We use \textit{relative mean square} convention for the norms of matrices and vectors. For example, the norm of a matrix $W\in\mathbb{R}^{n\times m}$ is computed as $\|W\|=\sqrt{\frac{1}{nm}\sum_{i=1}^n\sum_{j=1}^m W_{ij}^2}$.

\section{Learnable Multipliers}\label{sec:learnable_multipliers}
Consider a linear layer $y_i = \sum_j \overline{W}_{ij}x_j$, where $x_j \in \mathbb{R}^{d_{in}}$ and $y_i \in \mathbb{R}^{d_{out}}$ are the input and output features, $\overline{W}_{ij}\in\mathbb{R}^{d_{in}\times d_{out}}$ is the feature map matrix, and we have used index notation for vectors and matrices. The weight reparametrization amounts to using another matrix $W_{ij}$ and possibly additional weights to be learned by the optimization algorithm instead of the \emph{effective weight matrix} $\overline{W}_{ij}$. In this work, we reparametrize $\overline{W}_{ij}$ with either scalar multiplier $s$ or vector multipliers $r_i,c_j$. 

To escape the noise-WD equilibrium value \eqref{eq:equilibrium_norm_scaling} of the feature map matrix $\overline{W}_{ij}$, it is sufficient to add  

\begin{equation}\label{eq:scalar_multiplier}
    \hspace{-29mm}\text{Scalar Multiplier}: \qquad \overline{W}_{ij} = s W_{ij}, \quad s\in\mathbb{R}.
\end{equation}
Here, the learnable matrix weight $W_{ij}$ is still subject to the noise-WD equilibrium with the norm $\|W\|\propto\sqrt{\frac{\eta}{\lambda}}$. The scalar multiplier $s$ is supposed to learn freely so that the full matrix norm $\|\overline{W}\|=s\|W\|$ optimally adapts to a given data distribution. 

We make a step further and hypothesize that not only the norm of the whole matrix $\|W\|$, but also the norms of its individual rows $\|W_{i\,\bullet}\|$ and columns $\|W_{\bullet\,j}\|$ might also be stuck in the noise-WD equilibrium. Hence, we attach a learnable scale parameter to each row and column with 
\begin{equation}\label{eq:vector_multipliers}
    \text{Vector Multipliers}: \qquad \overline{W}_{ij} = r_i W_{ij} c_j, \quad r_i\in\mathbb{R}^{d_{out}}, \; c_j\in\mathbb{R}^{d_{in}}.
\end{equation}
As for the scalar case, $\|W\|$ is expected to have the equilibrium value, while each component of the learnable row $r_i$ and column $c_j$ multipliers is supposed to learn the respective optimal scale. 

It is instructive to relate the gradients of the reparametrized matrix $W_{ij}$ and the introduced multipliers $s,r_i,c_j$ to the gradients of the effective matrix $\overline{G}_{ij}=\frac{\partial \mathcal{L}}{\partial \overline{W}_{ij}}$, where $\mathcal{L}$ is the training loss. Direct application of the chain rule gives
\begin{equation}\label{eq:multipleir_gradients}
    \frac{\partial \mathcal{L}}{\partial W_{ij}}=r_ic_j\overline{G}_{ij}, \quad \frac{\partial \mathcal{L}}{\partial r_i} = \sum_j W_{ij}c_j\overline{G}_{ij},\quad \frac{\partial \mathcal{L}}{\partial c_j} = \sum_i r_iW_{ij}\overline{G}_{ij},\quad \frac{\partial \mathcal{L}}{\partial s} = \sum_{ij} W_{ij}\overline{G}_{ij}.
\end{equation}
We see that row/column multipliers accumulate the gradients across the respective column/row of the gradient matrix $\overline{G}_{ij}$, while scalar multiplier accumulates the gradients across the whole matrix. This extra averaging reduces the gradient noise level in the multipliers and intuitively explains why they do not experience noise-driven Brownian expansion that needs to be countered by weight decay. 

In fact, learnable multipliers are partially used in the nowadays standard Pre-LN architectures \citep{Xiong2020transformernorm} through RMSNorm learnable weights that can be viewed as column multipliers $c_j$ of the first linear layer in the block. While RMSNorm weights already provide scale adaptation to a part of the model, we argue that adding the multipliers to the remaining parts of the model yields further performance improvement.   

Finally, we note a natural idea of using a logarithmic scale for the learnable multipliers, for example, $s\rightarrow e^s$ in the scalar case. Such reparameterization can be beneficial when different multipliers tend to have both large and small scales, making it problematic to learn with uniformly scaled optimizer updates given by the multipliers' learning rate. However, in agreement with \citep{salimans2016weightnormalizationsimplereparameterization}, we observed that log-scale parametrization gives at most a slight performance advantage while posing stability issues we discuss in more detail in sec.~\ref{sec:symmetry}.

\paragraph{Model placement.} While vector multipliers \eqref{eq:vector_multipliers} are strictly more expressive than scalars \eqref{eq:scalar_multiplier}, using both row $r_i$ and column $c_j$ multipliers for all the matrix layers is clearly redundant. We previously mentioned that the column multiplier $c_j$ of the first linear layer in a block is equivalent to the RMSNorm weights of that block: using both is one example of such redundancy. Another example of redundancy is using both row multipliers for MLP up projection and column multipliers for down projection. As we explain in sec.\ref{sec:symmetry}, redundant multipliers give rise to a symmetry transformation in the model parameters that may lead to NaN values during training. We provide our recommended placement of multipliers for gated MLP, attention, and mamba2 \citep{dao2024transformersssmsgeneralizedmodels} blocks in sec.~\ref{sec:multipliers_placement}.  

\paragraph{Implementation.} For inference, learnable multipliers can be merged with their matrix $W_{ij}$ into the effective matrix $\overline{W}_{ij}$, and thus do not introduce any memory or latency overhead. 

During training, however, there are two distinct implementation strategies with different impacts on the training throughput. A simpler approach is to explicitly use the reparametrized expression \eqref{eq:scalar_multiplier},\eqref{eq:vector_multipliers} in the model forward pass, relying on the standard automatic differentiation and optimizer implementations to handle the update of the multipliers. In this case, the throughput is expected to drop at most by a couple of percent since the multiplier parameter count is tiny compared to the matrix layers. Yet, this drop can be further reduced by using effective matrices $\overline{W}_{ij}$ in the model's forward and backward pass, while manually handling the dynamics of multipliers and learnable matrix $W_{ij}$ on the optimizer level with the help of gradient relations \eqref{eq:multipleir_gradients}.    

\paragraph{Maximal update parametrization.} \citep{pmlr-v139-yang21c,yang2022tensorprogramsvtuning,yang2024tensor,dey2025dontlazycompletepenables} also use scalar reparameterization \eqref{eq:scalar_multiplier} but with non-learnable scale $s$ equipped with scaling rules w.r.t. model dimensions for predictable model size scaling. To maximize the performance, many pretrained language models additionally tune the multipliers on a smaller scale, and then transfer tuned multipliers to the target model scale \citep{dey2023cerebrasgptopencomputeoptimallanguage, hu2024minicpmunveilingpotentialsmall, zuo2025falconh1familyhybridheadlanguage}.

Our scalar multipliers \eqref{eq:scalar_multiplier} can be viewed as learnable version of $\mu$P multipliers, significantly affecting established $\mu$P workflows such as hyperparameter transfer. On the one hand, learnable $\mu$P multipliers no longer allow to enforcing $\mu$P scaling with model dimensions, and thus require a separate analysis of the model size scaling behavior. On the other hand, learnable multipliers reduce the need to perform compute-intensive tuning of the multipliers. As an example, \citep{zuo2025falconh1familyhybridheadlanguage} performed extensive tuning of 12 forward (weight reparametrization), 16 learning rate, and 7 weight decay multipliers. Our approach removes the need to tune both forward and weight decay multipliers that were responsible for the scale of model weights. 

\paragraph{Learnable multipliers in the literature.} Reparametrization of model weights with learnable multipliers was previously proposed in various deep learning contexts and for various reasons. Weight normalization \citep{salimans2016weightnormalizationsimplereparameterization} introduces multipliers as a replacement for batch normalization. Several works add learnable multipliers to residuals to enable stable training of very deep models \citep{Bachlechner2020ReZero,de2020batchnormalizationbiasesresidual,zhang2018residual,Huang2020tfixup, Nishida2024InitializationOL}. For transformer models, multipliers are used within parameter-efficient finetuning methods to increase expressivity \citep{liu2022fewshot, liu2024doraweightdecomposedlowrankadaptation,wang2024borabidimensionalweightdecomposedlowrank}, or together with extra normalization layers to address gradients mismatch along depth. Our perspective of using the multipliers to address the noise-WD equilibrium explains the mechanism by which they improve the performance and guides towards their comprehensive placement throughout the model architecture.

\begin{figure}[ht]               
  \centering
    \includegraphics[scale=0.43, clip, trim=9 13 10 10]{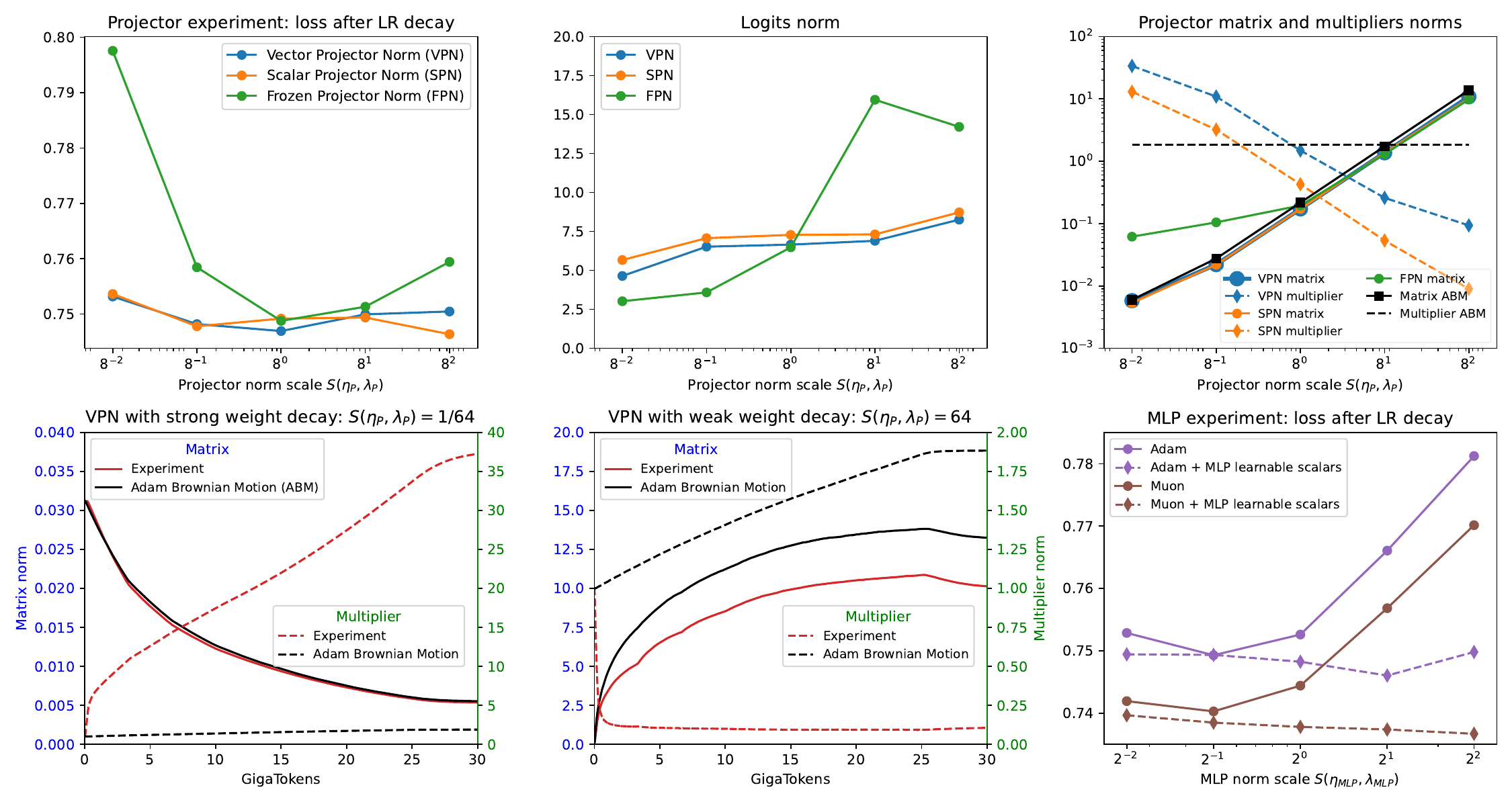}
  \caption{Projector and MLP scale \eqref{eq:equilibrium_norm_scaling} sweep experiments described in section~\ref{sec:whatlearned}. Norm scales $S(\eta_P,\lambda_P)$ and $S(\eta_{MLP},\lambda_{MLP})$ use relative values of learning rate and weight decay. \textit{(Top left):} the final loss of three projector norm configurations. \textit{(Top middle and right):} trajectories of projector $\|W\|$ and multiplier $\|\mathbf{c}\|$ norms during training are compared between the experiment and Adam Brownian Motion simulation. \textit{(Bottom left and middle):} logits and matrix/multipliers norms for the considered configuration. \textit{(Bottom right):} the final loss of three MLP experiment configurations.}
  \label{fig:projector}
\end{figure}

\section{What is learned by the multipliers?}\label{sec:whatlearned}

So far, the inability of matrix layers to learn data-dependent scales has been hypothesized but not experimentally validated. In this section, we support this hypothesis by providing different views on feature scales learned by the model.

We start by examining whether matrix layers can natively escape noise-WD equilibrium under significant loss optimization pressure. Under the scaling assumption \eqref{eq:equilibrium_norm_scaling}, varying the norm scale $S(\eta,\lambda)=\sqrt{\frac{\eta}{\lambda}}$ forces the model weights out of typical and presumably optimal scale. We perform two such tests for the LM head layer and projections of MLP block, with the results depicted in figure~\ref{fig:projector}. In these experiments, we vary $S(\eta,\lambda)$ while keeping the effective learning rate $\eta_\mathrm{eff}\equiv\sqrt{\eta\lambda}=\mathrm{const}$ \cite{zuo2025falconh1familyhybridheadlanguage} to isolate the effect of norm change from overall learning speed. 

\paragraph{Projector experiment.} Consider a standard final projector layer (LM head) $W_{ij}$ that maps normalized backbone features $\mathbf{x}, \operatorname{mean}\{x_j^2\}=1$ to probability logits $y_i=\sum_j W_{ij}c_jx_j$, where $c_j$ are learnable RMSNorm weights. We compare the following configurations differentiated by their multiplier type.
\begin{enumerate}
    \item \textbf{Frozen projector norm (FPN)}: $c_j\to1$. In this configuration, the logits scale is determined only by the scale of $W$ and its correlation with features $\mathbf{x}$.  
    \item \textbf{Scalar projector norm (SPN)}: $c_j\to s$ with learnable scalar multiplier $s$.  
    \item \textbf{Vector projector norm (VPN)}: a standard configuration with freely learnable $c_j$. 
\end{enumerate}
First, we observe in Figure~\ref{fig:projector} (top left) that the FPN configuration, where the logits scale relies entirely on $\|W\|$, suffers a clear performance drop at extreme values of the equilibrium projector norm $S(\eta_P,\lambda_P)=\sqrt{\frac{\eta_P}{\lambda_P}}$. In contrast, both the SPN and VPN configurations maintain stable performance. This performance gap can be traced to the logit norms (top middle), which remain stable (and presumably optimal) for the multiplier-equipped configurations, but vary significantly for FPN. Finally, we find that the norm $\|W\|$ for all the configurations (top right) indeed follows the noise-WD equilibrium scaling \eqref{eq:equilibrium_norm_scaling}, while scalar and vector multipliers are not restricted and adjust their scale to compensate for the large/small projector norm. The only exception to this equilibrium scaling is the FPN configuration at small equilibrium scales: the optimization pressure to maintain a non-vanishing logit norm is strong enough to pull the weights away from equilibrium.

To clearly identify the extent to which projector and multipliers follow a noise-driven dynamics, we compare their norms with \emph{Adam Brownian Motion} (ABM). To simulate pure noise-determined trajectories, we generate a sequence of zero mean i.i.d. gradients $g_t\sim\mathcal{N}(0,1)$ and use them in the AdamW optimizer update with the same $\lambda$ and schedule of $\eta$ as was used in the training. On figure~\ref{fig:projector} (bottom left and middle), we observe that ABM and experimental trajectories of $\|W\|$ fully coincide for strong WD and are quite close for weak WD, while experimental multipliers trajectories show no resemblance with their ABM version. The same applies to final norm values on figure~\ref{fig:projector} (top right). 

\paragraph{MLP experiment.} Next, we design a similar experiment for residual MLP blocks, where we vary the norm scale $S(\eta_{MLP},\lambda_{MLP})$ only for the 3 matrix layers of the gated MLP block. To test the ability of MLP matrix layers to adapt their scale we consider two configurations:
\begin{enumerate}
    \item Frozen RMSnorm weights across all backbone layers to restrict the scale adaption ability of all the backbone blocks. 
    \item Same as above, but with scalar learnable multipliers added to MLP block to compensate for varied norm scale $S(\eta_{MLP},\lambda_{MLP})$ of its matrix layers. 
\end{enumerate}
The loss behavior of these configurations is depicted on figure~\ref{fig:projector} (bottom right), where we observe that configuration without learnable multipliers suffers from loss degradation at larger norm scales $S(\eta_{MLP},\lambda_{MLP})$ of MLP matrices. We attribute this degradation to output's magnitude mismatch between MLP and attention/SSM blocks, and investigate it in more details in section~\ref{sec:MLP_exp}. Importantly, we perform MLP experiment for both Adam and Muon optimizers and observe essentially identical behavior between the two optimizers. This suggest that noise-WD equilibrium trap is not a specific property of Adam but a general phenomenon persisting across different parameter update rules. 

\subsection{Features scale diversity}
Having confirmed the ability of multipliers to learn the scale that was fixed in matrix-only architecture configurations, we proceed with an investigation of new features that are learned when the model scales are freed by adding the multipliers.

\begin{figure}[t]               
  \centering
    \includegraphics[scale=0.42, clip, trim=10 2 10 10]{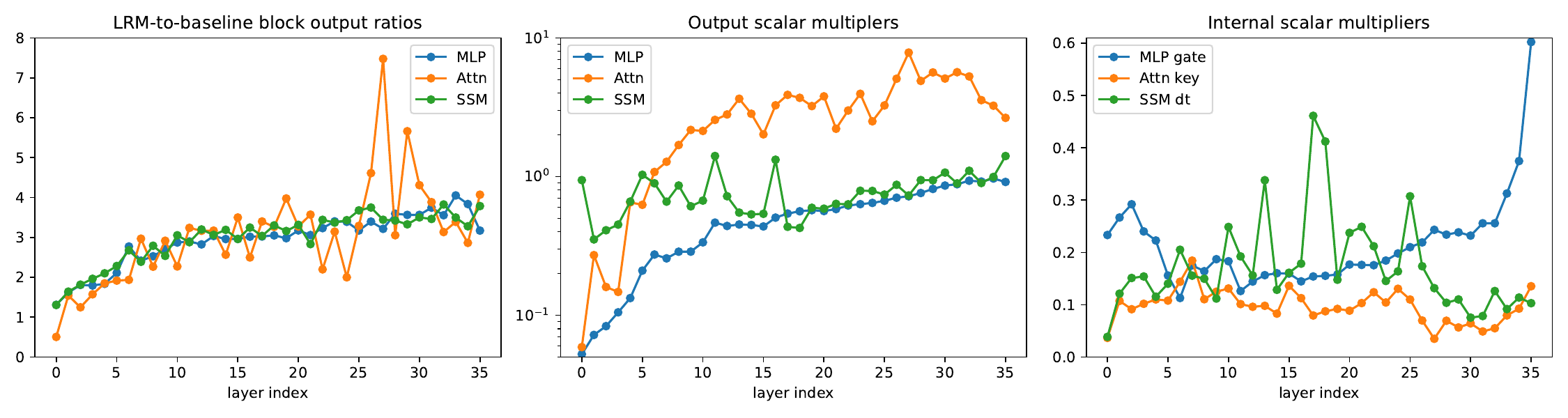}
  \caption{\textit{(Left):} a ratio of residual block output norms of the configuration with scalar multipliers to the configuration without. \textit{(Middle):} values of scalar multipliers at the end of each block. \textit{(Right):} values of internal scalar multipliers at selected locations of each block.}
  \label{fig:depth_wise_SM}
\end{figure}

\paragraph{Depth-wise scales.} We add scalar multipliers to all matrix layers in the model and measure the norms of residual blocks output across the model's depth. The results are depicted in figure~\ref{fig:depth_wise_SM}.  

For block outputs we observe an overall increasing trend towards later layers, suggesting that larger contribution of these layers to the residual could be beneficial for loss optimization but could not be fully learned by classical architecture without multipliers. Moreover, we observe that attention layers in the second half of the model have significantly different scales that are further amplified by multipliers values (left and middle subfigures). Additionally, dt projection that controls memorization and forgetting in the SSM block show significant variation across the layers, suggesting specializing of those layers in varying temporal scales.       

\begin{figure}[t]               
  \centering
    \includegraphics[scale=0.53]{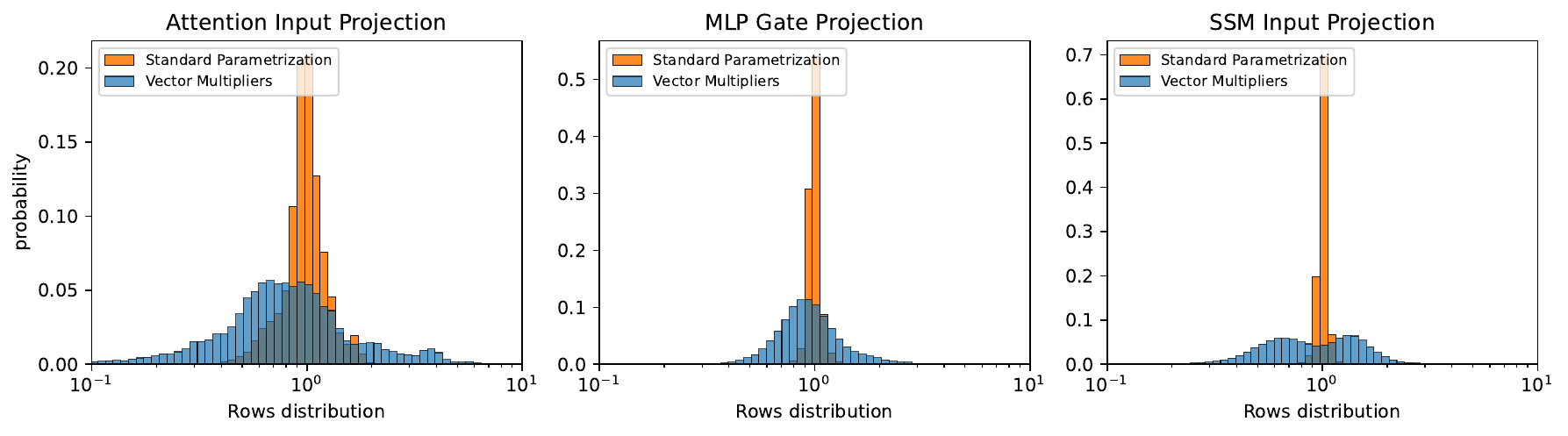}
  \caption{Distributions of row norms $\|W_{i\,\bullet}\|$ of attention/SSM input  and MLP gate projections, which correspond to internal features of these blocks. We collect the norms across all the model layers while normalizing norm values of each layer by their mean to align the scale of different layers and focus on within-layer distribution.}

  \label{fig:output_features_density}
\end{figure}

\paragraph{Width-wise scales.} 
To motivate vector multipliers \eqref{eq:vector_multipliers} in section~\ref{sec:learnable_multipliers}, we assumed that row $\|W_{i\,\bullet}\|$ and column $\|W_{\bullet\,j}\|$ norms are also subject to noise-WD equilibrium. Now, we confirm this assumption on figure~\ref{fig:output_features_density} by measuring the distribution of row (output feature) norms of the effective layer matrices $\overline{W}$ for configurations with and without learnable vector multipliers. 

We select attention/SSM input and MLP gate projections, whose outputs go into essential non-linear transformations of those layers, and hence very sensitive to the scale. Yet, the configuration without multipliers show a very narrow distribution implying low scale diversity of internal features in these layers. Adding learnable vector multipliers significantly broadens the norm distribution, suggesting that a larger diversity of internal feature scales in these blocks is beneficial for the model.       

\section{Aspects of multiplier training dynamics}\label{sec:details}

\subsection{Symmetries}\label{sec:symmetry}

\begin{figure}[t]               
  \centering
  \begin{subfigure}[b]{0.4 \linewidth}
    \centering
    \includegraphics[width=\linewidth]{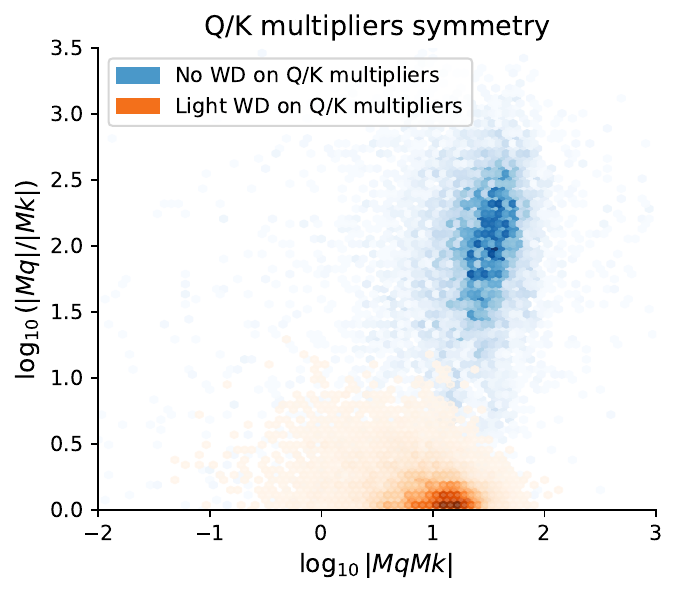}
    \caption{}
    \label{fig:qk_symmetry}
  \end{subfigure}
  \begin{subfigure}[b]{0.4\linewidth}
    \centering
    \includegraphics[width=\linewidth]{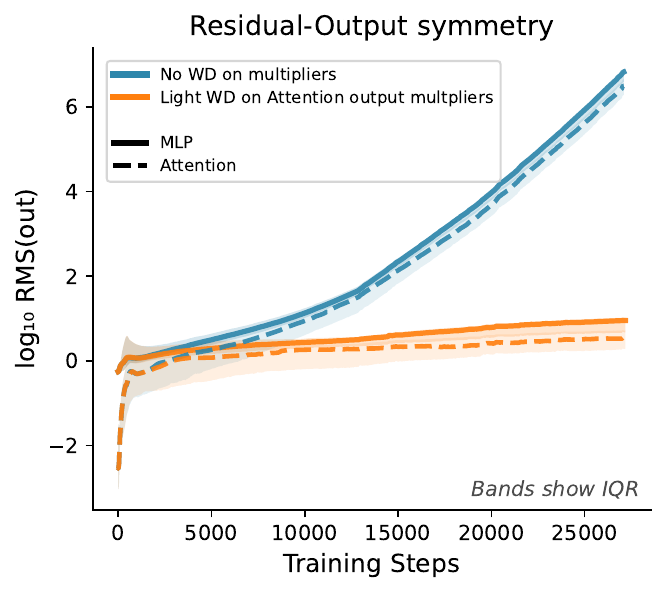}
    \caption{}
    \label{fig:residual_symmetry}
  \end{subfigure}

  \caption{
    Effect of light WD on (a) Q/K multiplier symmetry and (b) residual output symmetry. Light WD (orange) suppresses the drift and unbounded norm growth observed in the baseline without WD (blue).
}
  \label{fig:ng_ds_combined}
\end{figure}

Unlike weight matrices which require weight decay to bound their scale, learnable multipliers seem to not require any scale control due to their ability to freely learn the optimal scale. However, the multipliers are vulnerable to a different source of scale instability: architecture symmetries. These symmetries represent scaling transformation in the parameter space of the model that does not change final model's output \citep{dinh2017sharp}. While harmless in exact arithmetic, this drift introduces significant instability under low-precision formats like \texttt{bfloat16}, as it increases quantization error and degrades gradient estimates \citep{micikevicius2017mixed,Angelo2024whyweightdecay,bloom2022}. We highlight two symmetries common in language model architectures:
\begin{itemize}
  \item \textbf{Multiplicative symmetry.} When two learnable factors $a$ and $b$ appear only through their product $ab$, the reparameterization $(a,b)\mapsto (sa, s^{-1}b)$ leaves the forward map unchanged for any $s\neq0$. As a result, $\lVert a\rVert$ can grow while $\lVert b\rVert$ shrinks. An example of that is the product of queries $Q$ and keys $K$ in the attention computation. Figure~\ref{fig:qk_symmetry} illustrates this: for the baseline without intervention (blue), the product of the $Q/K$ multipliers (x-axis) remains stable while their scale ratio (y-axis) drifts significantly.
\item \textbf{Normalization symmetry.}
Assume the model uses a quantity \(c\) only through its normalized version
\(\hat c := c/\|c\|_{\mathrm{rms}}\) (e.g., the activation before the final head projection is rescaled to unit RMS: Only the relative magnitudes of the residuals producing that activation matters).
Then for any \(s>0\), the rescaling \(c\mapsto s c\) leaves \(\hat c\), and hence the model output unchanged.
Consequently, the residuals norms can grow without bound while the final output remains identical. This unbounded growth is demonstrated in Figure~\ref{fig:residual_symmetry}, where the RMS of the residual outputs (blue lines) increases over training steps when no weight decay is applied.
\end{itemize}

In standard architecture without learnable multipliers, these symmetries are essentially fixed by equilibrium norms of the weight matrices. When we first added learnable multipliers, we have a drift along symmetry directions, leading to large activations, and, eventually, divergence or NaN values. 

There are several possible approaches to address the symmetry-induced instabilities. One may manually rescale weights along the symmetry directions back to ``normal'' values or remove all the multipliers subject to the symmetry transformations, with specific configuration provided in section \ref{sec:multipliers_placement}. The former requires extensive architecture-specific engineering, and we have found the latter to surprisingly reduce final model performance. We converged to applying a small weight decay $\lambda_{lrm} = 2\times10^{-3}$ to multipliers as a simple yet effective solution to handle the symmetries.     

\begin{table}[t]
\centering
\footnotesize
\begin{tabular*}{\textwidth}{l @{\extracolsep{\fill}} *{5}{c}}
Scaling recipe & LR $\eta$ & WD $\lambda$ & Multiplier $s$ & $\|\mathbf{y}\|$ & $\frac{\|\Delta\mathbf{y}\|}{\|\mathbf{y}\|}$ \\
\cmidrule(r){1-1} \cmidrule{2-4} \cmidrule(l){5-6}
Standard through LR & $d^{-1}$ & $\mathrm{const}$ & $\mathrm{const}$ & $\color{red}d^{\frac{1}{2}}$ & $\color{red}d^{-\frac{1}{2}}$ \\
\midrule
Through LR \& WD & $d^{-1}$ & $d$ & $\mathrm{const}$ & $\color{green}\mathrm{const}$ & $\color{green}\mathrm{const}$ \\
\midrule
Through multiplier & $\mathrm{const}$ & $\mathrm{const}$ & $d^{-1}$ & $\color{green}\mathrm{const}$ & $\color{green}\mathrm{const}$ \\
\midrule

\end{tabular*}

\vspace{2mm}

\caption{Scaling of the output norm $\|\mathbf{y}\|$ of a linear layer $\mathbf{y}=sW\mathbf{x}, \; \mathbf{x}\in\mathbb{R}^{d}, s\in\mathbb{R}$, and relative norm $\frac{\|\Delta\mathbf{y}\|}{\|\mathbf{y}\|}$ of the update $\Delta\mathbf{y}=\Delta W\mathbf{x}$ with width $d$. 
The standard LR-only scaling recipe is taken from \cite{yang2022tensorprogramsvtuning,dey2023cerebrasgptopencomputeoptimallanguage,hu2024minicpmunveilingpotentialsmall}, its adjustment to include WD is given in \cite{kosson2025weightdecaymattermup}, and an alternative version that scales multipliers was used in \cite{zuo2025falconh1familyhybridheadlanguage}.} 
\label{tab:mup_scaling}
\end{table}

\subsection{Scaling with model width}\label{sec:width_scaling}
\begin{figure}[t]               
  \centering
  \includegraphics[scale=0.53, clip, trim=10 10 10 10]{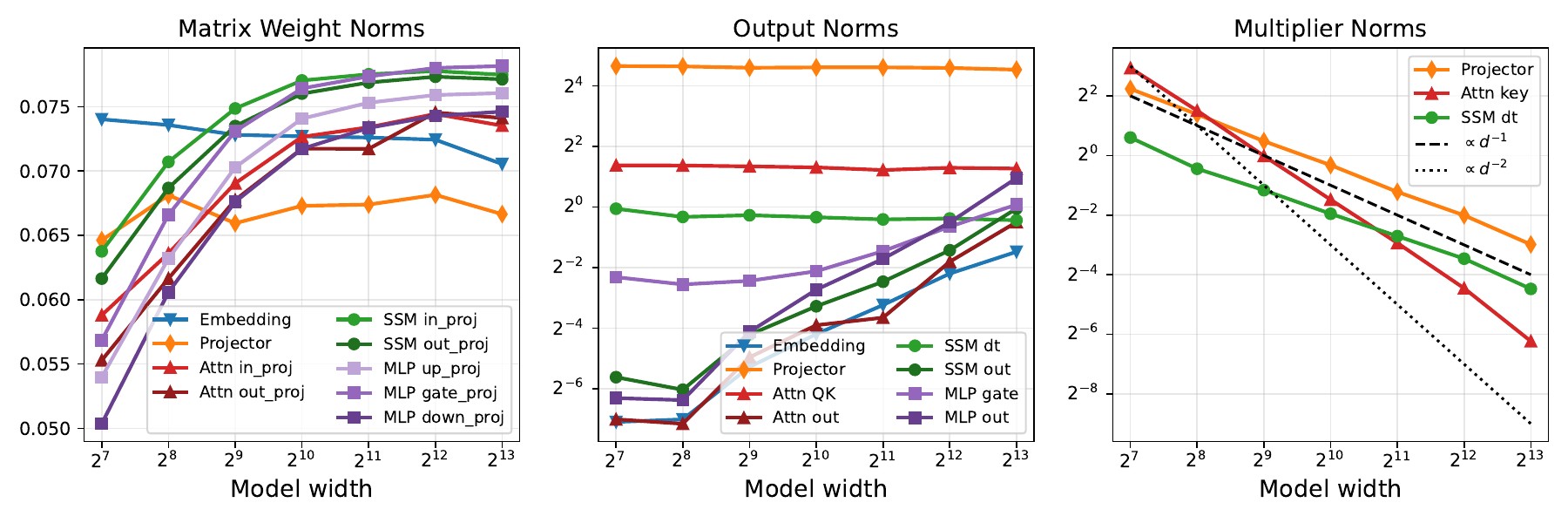}
  \caption{The width scaling of the norms of linear layers matrices, various activations throughout the model, and scalar multipliers attached to the selected model activations. We use geometric average to aggregate the values across the layers. Other experimental details can be found in section \ref{sec:details}, and we provide time evolution of selected output norms on figure \ref{fig:width_mup_out_norms}.}
  \label{fig:multipliers_width_mup}
\end{figure}

When the size of a model is scaled, a natural requirement is to keep constant the magnitude of model activations for stable and performant forward pass, while efficient feature learning requires keeping constant the magnitude of activation updates (see, for example, desideratum 1 of \cite{yang2024spectralconditionfeaturelearning}). While $\mu P$ satisfies these requirements with certain scaling rules of forward multipliers and/or learning rate, both the presence of equilibrium \eqref{eq:equilibrium_norm_scaling} and the addition of learnable multipliers prompt revisiting of $\mu P$ scaling rules. In this section, we briefly explore the above questions via an experiment where we (i) use scalar learnable multipliers \eqref{eq:scalar_multiplier} (ii) vary model width $d$ (iii) keep learning rate $\eta$ and weight decay $\lambda$ fixed across widths $d$. The results are depicted on figure~\ref{fig:multipliers_width_mup}.

\paragraph{Equilibrium matrix norms.} First, we observe on figure~\ref{fig:multipliers_width_mup} (left) that the norms of matrix layers stay almost constant across considered model widths. This suggests that the equilibrium norm does not scale with model size, for example, via increased noise level at larger sizes. Such conclusion is also consistent with width-agnostic Adam Brownian Motion (ABM) model that accurately describes equilibrium norm in our projector experiment in figure~\ref{fig:projector}. Note that we still see a slight growth of the norms at very small widths. One interpretation of this growth is that the noisy regime described by \eqref{eq:equilibrium_norm_scaling} requires a low enough ratio of signal (e.g. measured by \# of tokens in the batch) to model capacity (e.g. measured by \# of model parameters).

\paragraph{Implications for $\mu$P scaling rules.} 
Let us now highlight the effect of equilibrium norm on LR-only $\mu$P scaling recipe typically used in documented LLM applications, and outlined in the top row of table \ref{tab:mup_scaling}. Consider a linear layer $\mathbf{y}=W\mathbf{x}, \; \mathbf{x}\in\mathbb{R}^{d}$ without learnable multipliers.
Then, a width-independent equilibrium \eqref{eq:equilibrium_norm_scaling} breaks the optimal hyperparameter transfer of this scaling rule, leading to exploding activation $\|\mathbf{y}\|\propto d^{\frac{1}{2}}$ and vanishing relative update strength $\tfrac{\|\Delta\mathbf{y}\|}{\|\mathbf{y}\|}\propto d^{-\frac{1}{2}}$. Indeed, assuming normalized input $\|\mathbf{x}\|=\mathrm{const}$, fixed alignment between $\mathbf{x}$ and $W$, and equilibrium norm $\|W\|\propto\sqrt{\frac{\eta}{\lambda}}$, we have $\|\mathbf{y}\|\propto d \|W\| \|\mathbf{x}\|\propto d \times \sqrt{\frac{\eta}{\lambda}} \propto d^{\frac{1}{2}}$. Similarly, for the relative updates $\frac{\|\Delta \mathbf{y}\|}{\|\mathbf{y}\|}\propto\frac{d\|\Delta W\|\|\mathbf{x}\|}{d\| W\|\|\mathbf{x}\|}=\frac{\|\Delta W\|}{\|W\|}\propto\frac{\eta}{\sqrt{\eta/\lambda}}=\sqrt{\eta\lambda}\propto d^{-\frac{1}{2}}$, where we have assumed a scale-free optimizer with $\|\Delta W\|\propto \eta$, and fixed alignment between $\Delta W$ and $\mathbf{x}$. 

The breakdown of LR-only recipe demonstrated above actually originates from omitting weight decay scaling rather than from a shortcoming of underlying $\mu P$ foundations. To have a maximal feature learning infinite width limit, \cite{yang2023tensorprogramsivbadaptive} notes that the total weight decay coefficient $\eta\lambda$ needs to be fixed when scaling width $d$. This requirement already fixes the effective learning rate $\eta_\mathrm{eff}=\sqrt{\eta\lambda}$ which governs relative magnitude of updates $\frac{\|\Delta \mathbf{y}\|}{\|\mathbf{y}\|}\propto \eta_\mathrm{eff}$. Then, constant magnitude of activations $\|\mathbf{y}\|$ can be achieved by either adding weight decay scaling $\lambda\propto d$, or, by reparametrizing the layer as $\mathbf{y}=sW\mathbf{x}$ and moving the scaling to the multiplier: $\eta,\lambda=\mathrm{const}, \; s\propto d^{-1}$. The second approach naturally connects to the learnable multipliers and allows us to compare the learned and $\mu P$-scaled values of $s$, that we discuss below.     
 
\paragraph{Learnable multipliers scaling.} Now, we return to our experiment with scalar learnable multipliers to check if the right width scaling is learned automatically by the multipliers. As the relative scale of activation update $\frac{\|\Delta \mathbf{y}\|}{\|\mathbf{y}\|}$ induced by matrix update $\Delta W$ is independent from multiplier $s$, it is sufficient to look only at the scale of activations $\|\mathbf{y}\|$. 

Indeed, on figure \ref{fig:multipliers_width_mup} (middle) we observe stable activation norms of projector (model's logits) and selected internal activations: product of the norms of attention keys and values which governing attention logits scale and SSM $\texttt{dt}$ activation that governs forgetting/memorization of the SSM hidden state. While the norms of embeddings and residual blocks outputs jointly grow with width, we note that they are, in a sense, ambiguous due to the residual normalization symmetry. The MLP gate activation seem to fall somewhere in between: formally, it should have fixed scale due to $\operatorname{SiLU}(\cdot)$ non-linearity applied to it, but as $\operatorname{SiLU}(\cdot)$ asymptotically reduces to homogeneous $\operatorname{ReLU}(\cdot)$, the MLP gate is subject to the same residual normalization symmetry. Overall, stable projector, SSM $\texttt{dt}$ and attention \texttt{QK} norms confirms that LRMs adjust to width and learn the required scale.     

On figure \ref{fig:width_mup_out_norms} (right) we look at the actual value of the learned multipliers and compare it to expected $\mu P$ scaling: $d^{-1}$ for projector and SSM \texttt{dt}, and $d^{-2}$ for attention \texttt{QK}, where a factor of $d$ comes from each of \texttt{Q} and \texttt{K}. We observe that all the three considered multipliers scale only a slightly slower than the predicted $d^{-1}$ and $d^{-2}$ trends. To locate the source of this mismatch, let us define the alignment $\alpha$ between $W$ and $\mathbf{x}$ such that $\|\mathbf{y}\|=s d \alpha\|W\|\|\mathbf{x}\|$, and $\alpha\propto \mathrm{const}(d)$ corresponds to the standard $\mu P$ limit assumption. As the absence of width scaling of $\|W\|$ and $\|\mathbf{y}\|$ was confirmed independently, and $\|\mathbf{x}\|=1$ by architecture design, we conclude that $\alpha \propto (sd)^{-1}\propto d^{a}, a<0$. This decay of alignment $\alpha$ between the input features and model weights contradicts the established $\mu P$ regime and prompts further investigation beyond the scope of current work.      

\subsection{Gradient clipping} 
\begin{figure}[t]               
  \centering
    \includegraphics[scale=0.45]{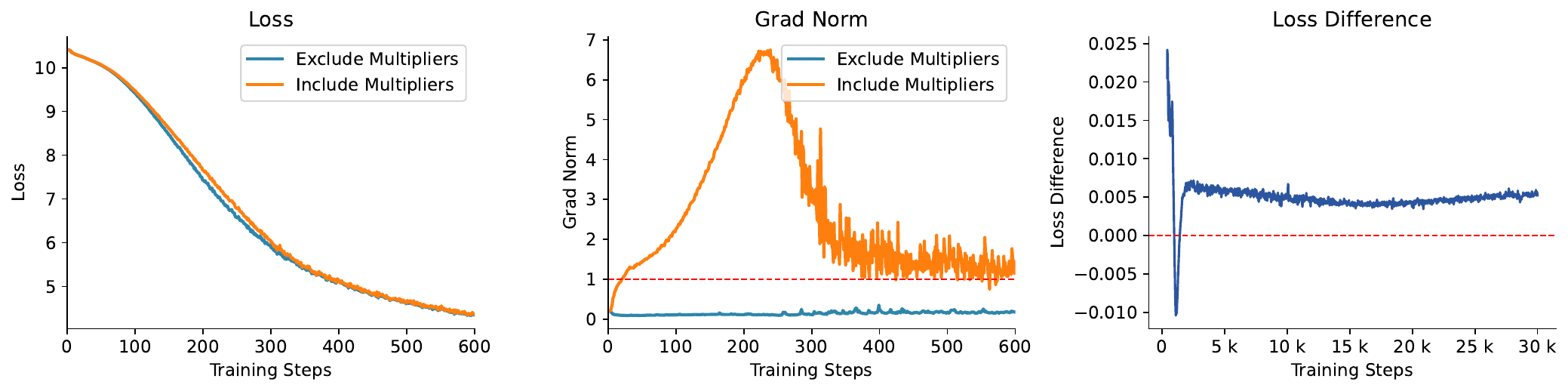}
  \caption{
    Training dynamics when excluding (blue) vs. including (orange) multiplier gradients from the global clip norm: (Left) initial loss, (Middle) initial gradient norm, and (Right) the long-term loss difference.
}
  \label{fig:grad_clip}
\end{figure}

Clipping  $\ell_2$-norm of the gradients is a standard approach aimed to improve training stability \citep{pascanu2013difficulty}. We also followed this practice and applied and applied clipping at the value of $1$. Contradictory to the intuition developed in this work, we have observed from none to negative impact of adding scalar learnable multiplier in our initial experiments. The unsatisfactory performance was revealed to be related gradient clipping as demonstrated on figure \ref{fig:grad_clip} (middle): the run with multipliers had large gradient norms in the initial stages of the training that were clipped, while the gradient norms excluding the contribution from multipliers were significantly below the clipping threshold. This forces an overly aggressive clipping factor that unnecessarily shrinks the updates for \emph{all} parameters. Excluding multipliers from grad norm computation mitigated the issue and resulted in significantly better performance, as shown in figure \ref{fig:grad_clip} (right).

\subsection{Learning the projector scale}\label{sec:projector}
The scale of the final output layer of a neural network has a strong impact on its feature learning ability: large scale leads to so called ``lazy training'' regime while smaller scales force backbone feature to update significantly to reach the required change in model outputs \cite{Chizat2019_lazy, pmlr-v139-yang21c}. Attaching learnable multiplier to the projector (LM head) layer might induce transition between feature learning and lazy regimes, and, therefore, requires careful consideration. 

Starting with a standard configuration with final RMSNorm weights acting as projector column multipliers $c_j$, and we first added row multipliers $r_i$ that allow to directly adjust the scale of each individual logit. The addition of $r_i$ resulted in performance degradation that can be explained within lazy training paradigm: direct learning of individual logits creates a shortcut to quickly fit marginal token distribution without meaningfully updating internal model features. Next, we tried to also remove column multipliers $c_j$ while sweeping over fixed scalar multiplier $s$ to further enhance feature learning strength \cite{atanasov2025the}, but did not observe an expected improvement in performance and leave further investigation for future work.        

\section{Results}\label{sec:results}
\subsection{Multiplier tuning ablation} 
Following \cite{zuo2025falconh1familyhybridheadlanguage}, $\mu$P model size scaling naturally suggest to tune (i) learning rate, (ii) weight decay and (iii) forward multipliers in order to improve model performance. Our approach automatically removes the need to tune forward multipliers as they become learnable. The need for WD multipliers as a mean to control the weight norms also becomes questionable. To jointly clarify the effects of tuning each hyperparameter type and learning vector multipliers \eqref{eq:vector_multipliers}, we consider the following 4 tuning configurations with progressive tuning cost
\begin{enumerate}
    \item \textbf{NONE}: no tuned multipliers (i.e. all set to $1$).
    \item \textbf{LR}: only LR multipliers are tuned while forward and WD multipliers are set to $1$.
    \item \textbf{LRWD}: LR and WD multipliers are tuned while forward multipliers are set to $1$.
    \item \textbf{FULL}: All multipliers are tuned.
\end{enumerate}

\begin{figure}[t]               
  \centering
    \includegraphics[scale=0.46, clip, trim=11 12 10 10]{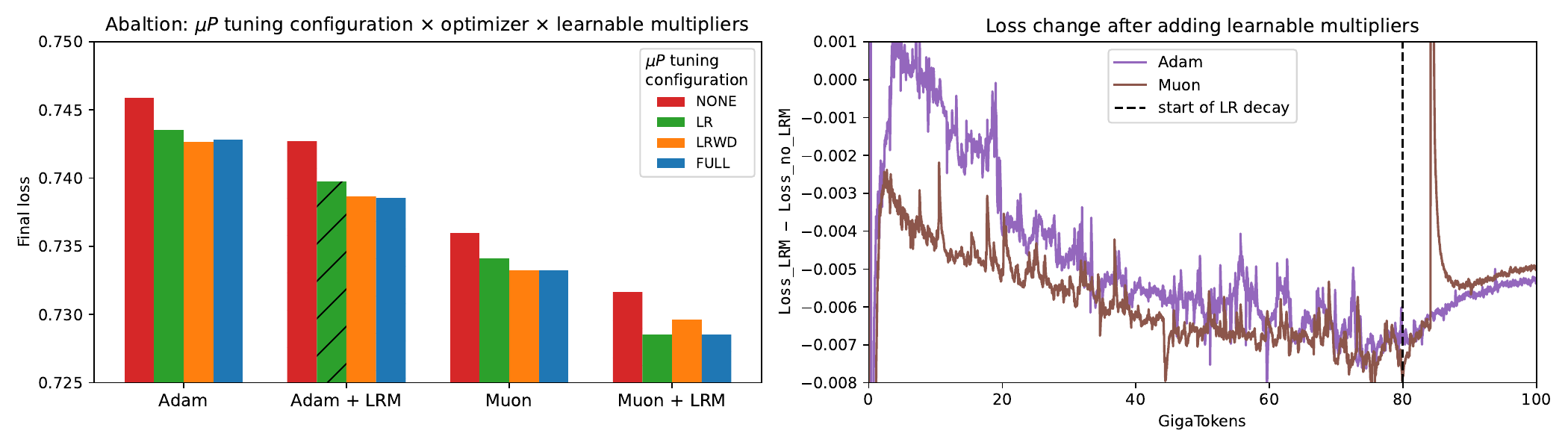}
  \caption{\textit{(Left)}: Loss values after LR decay for 4 multiplier tuning configurations. Each configuration is trained both with standard parametrization and with learnable vector multipliers, marked with \texttt{+LRM}. \textit{(Right)}: loss difference between the runs with and without learnable multipliers.}
  \label{fig:init_ablation_plus_longrun}
\end{figure}

We perform this ablation for both Adam and Muon optimizers. The tuned values are borrowed from \cite{zuo2025falconh1familyhybridheadlanguage}; in the case where forward multipliers are learnable, the tuned values are used as initialization of the multipliers. The final loss values are depicted in figure~\ref{fig:init_ablation_plus_longrun} (left). We observe a consistent tiering of results for both optimizers: (i) non-learnable and fully not tuned is clearly the worst; (ii) non-learnable configurations with at least learning rate tuning are close to learnable multipliers without any tuning and provide moderate performance boost; (iii) learnable multipliers with learning rate tuning are close to each other and provide the most performance boost. Crucially, this hierarchy holds for Muon as well, confirming that learnable multipliers provide a generalized benefit that complements the optimizer's specific update rule.

Such grouping prompts several observations. First, learnable multipliers always improve performance of its parent configuration. Second, learnable multipliers do not require tuned initialization or WD multipliers, aligned with expectation that multipliers automatically converge to the optimal scale. Thirdly, tuning the LR multipliers appears essential and, by design, cannot be covered with learnable forward multipliers. Lastly, we note surprisingly absent effect of tuning forward an WD for non learnable configurations. We speculate that this lost performance boost may come from discrepancy between our training data mix and the mix used for tuning. This highlights a brittle and narrow nature of the tuning in contrast to robust data adaptation of learnable approach.  

In figure~\ref{fig:init_ablation_plus_longrun} (right) we depict the loss difference between the run with and without learnable multipliers. Learnable multipliers develop a loss gap that continues to slowly grow throughout the constant LR stage while slightly shrinking during LR decay. The latter shrinking may be explained from noise point of view: LR decay reduces the noise level in matrix layers, resulting in a less restrictive noise-WD equilibrium with an ability to partially learn the scale. The increase in the loss gap during training supports that the role of learnable multipliers is to increase feature scale diversity and thus arrive at asymptotically better performance, in contrast to simply ``speeding up'' the training.

\begin{table}[t]
\centering
\footnotesize

\begin{tabular*}{\textwidth}{l @{\extracolsep{\fill}} *{8}{c}}
\toprule
\textbf{Optimizer} & \textbf{Hellaswag} & \textbf{ARC-C} & \textbf{MMLU} & \textbf{MMLU-PRO} & \textbf{BBH} & \textbf{GSM8K} & \textbf{\makecell{MATH \\ lvl5}} & \textbf{Average} \\
\cmidrule(r){1-1} \cmidrule(l){2-9}
Adam & 48.91 & 38.70 & 44.18 & 18.26 & 9.70 & 48.35 & 7.52 & 30.80 \\
\midrule
\makecell[l]{Adam+LRM} &
    \makecell{49.89 \\ \scriptsize{\textcolor{GoodGain}{(+0.98)}}} &
    \makecell{38.73 \\ \scriptsize{\textcolor{GoodGain}{(+0.03)}}} &
    \makecell{45.33 \\ \scriptsize{\textcolor{GoodGain}{(+1.15)}}} &
    \makecell{\textbf{19.92} \\ \scriptsize{\textcolor{GoodGain}{(+1.66)}}} &
    \makecell{12.03 \\ \scriptsize{\textcolor{GoodGain}{(+2.33)}}} &
    \makecell{49.10 \\ \scriptsize{\textcolor{GoodGain}{(+0.75)}}} &
    \makecell{9.07 \\ \scriptsize{\textcolor{GoodGain}{(+1.55)}}} &
    \makecell{32.01 \\ \scriptsize{\textcolor{GoodGain}{(+1.21)}}} \\
\midrule
Muon & 50.31 & 39.04 & \textbf{47.98} & 18.96 & 10.13 & 48.02 & 8.69 & 31.88 \\
\midrule
\makecell[l]{Muon+LRM} &
    \makecell{\textbf{50.56} \\ \scriptsize{\textcolor{GoodGain}{(+0.25)}}} &
    \makecell{\textbf{39.39} \\ \scriptsize{\textcolor{GoodGain}{(+0.35)}}} &
    \makecell{47.96 \\ \scriptsize{\textcolor{BadGain}{(-0.02)}}} &
    \makecell{19.32 \\ \scriptsize{\textcolor{GoodGain}{(+0.36)}}} &
    \makecell{\textbf{13.52} \\ \scriptsize{\textcolor{GoodGain}{(+3.39)}}} &
    \makecell{\textbf{50.63} \\ \scriptsize{\textcolor{GoodGain}{(+2.61)}}} &
    \makecell{\textbf{9.49} \\ \scriptsize{\textcolor{GoodGain}{(+0.80)}}} &
    \makecell{\textbf{32.98} \\ \scriptsize{\textcolor{GoodGain}{(+1.10)}}} \\
\bottomrule
\end{tabular*}

\vspace{2mm}

\caption{Performance comparison of Baseline vs. Learnable multipliers settings. Gains are in \textcolor{GoodGain}{green}, losses are in \textcolor{BadGain}{red}, and the best score per benchmark is in \textbf{bold}. All values are percentages (\%). The reported evaluation score are obtained by averaging over checkpoints obtained at an additional 40GT of training after the LR decay. The full evaluation trajectories are reported in figure \ref{fig:eval_trajectories}.} 
\label{tab:evals}
\end{table}

\subsection{Long training validation} 
As a final validation of our approach, we perform a longer, 200GT duration run for configuration with all tuned multipliers, identical to that of \texttt{Falcon-H1-0.5B}, and its version with learnable vector multipliers. This duration roughly corresponds to $\times20$ of Chinchilla compute-optimal duration, and, therefore, reasonably mimics real pretraining settings. 

Table~\ref{tab:evals} details the downstream performance. The full evaluation trajectories for these runs are reported in Figure~\ref{fig:eval_trajectories} in the Appendix. While Muon itself is a stronger baseline than Adam (31.88\% vs. 30.80\% average benchmark score), applying learnable multipliers provide equal boost other the respective baseline: 1.21\% for Adam and 1.10\% for Muon.  This suggests that the ``noise-WD equilibrium'' trap is a general phenomenon affecting various optimizers, and learnable multipliers are a universal solution to escape it.

We observe consistent improvement from LRM across the capabilities board in table ~\ref{tab:evals}. Yet, one may notice a general trend: modest improvement for knowledge related benchmarks (ARC-C, MMLU), and a much more significant boost for reasoning related benchmarks (BBH, MATH lvl5, GSM8K). This suggest an uneven impact of learnable multipliers on different model capabilities, though a comprehensive investigation is required to confirm this hypothesis.     

\section{Conclusion and discussion}
We have shown that the equilibrium between noise expansion and weight decay, experienced by matrix layers, significantly reduces the scale diversity of model internal representations. To address this problem, we have added learnable multipliers to suitable locations in the model architecture. These multipliers adjust the fixed scale of matrix layers to the underlying training data and ensure richer representation scales. We validate learnable multipliers in a realistic setting of a long language model pre-training run and observe a sizable improvement. Thus, we conclude that reparameterizing matrix weights with learnable multipliers is a universal path to improving the pretraining performance without any sacrifice in the inference speed or memory cost.

Yet, many questions are left open. In this work, we heavily rely on clear distinction between matrix and scalar/vector weights: dynamics of matrices is noise-dominated and requires WD to counter Brownian expansion, while dynamics of scalar and vectors seem to be signal-dominated, removing the need of WD and ensuring the ability to freely learn the optimal scale. However, signal-to-noise ratio in the weight gradients forms a continuous spectrum, suggesting the existing of a certain criteria of when a parameter tensor acquires scale-adaptation ability. Hence, an interesting direction for future work is to mechanistically understand the difference between matrix and scalar/vector dynamics, find an empirically measurable indicator of the noise level, or build a minimal mathematical model exhibiting both training regimes. 

Next set of question is related to developing a complete set of scaling rules, generalizing classical $\mu$P scaling to the presence of learnable multipliers. For example, should we scale LR and WD (which constraints symmetries) of multipliers with model size? Or, does application of learnable multipliers automatically ensures maximal feature learning in infinite-width limit without manual scaling rules required in classical $\mu$P \cite{yang2023tensorprogramsivbadaptive}?

It is practically relevant to further investigate the the relation between learnable multipliers and the difference in improvement it provides to different capabilities we have preliminary seen in table \ref{tab:evals}. A interesting hypothesis to explore is whether learned multipliers enhance only a certain types of circuits learned by the model \cite{elhage2021mathematical}. How training stability and performance improvement of LRMs scale with model size is another practical question.  

Finally, we may rephrase the improvement of learnable multipliers over standard architecture in a more general way: standard training procedure has internal flaws preventing converging of the model to the global minimum of population loss for a given data distribution and model architecture, even at asymptotically long training; these flaws must be explicitly addressed to access loss values closer to global optimum. The unlearned matrix scale, corrected by learnable multipliers, is one example of such flaw and its correction. It is an open question whether there are other flaws such kind and whether they can be corrected. For example, are there other parts of parameter matrices apart from row and column norms that are not learned automatically?    

\bibliography{iclr2026_conference}
\bibliographystyle{iclr2026_conference}

\newpage
\appendix

\section{Experiment settings}
We perform all our experiments with \texttt{Falcon-H1-0.5B} architecture \citep{zuo2025falconh1familyhybridheadlanguage}. The main reason for this choice is the availability of 35 extensively tuned $\mu$P multipliers that serve as a strong baseline for our learnable multipliers. The 0.5B model scale provides a reasonable tradeoff between the model's ability and the computational cost of running multiple experiments. Finally, we rely on the available training infrastructure for hybrid attention-SSM models to reduce infrastructure implementations and focus on the multiplier-related aspects. 

The training duration for most of the experiments is fixed to 30 GT, comprised of 25GT of the constant learning rate stage and 5GT of exponential LR decay with $\times 8$ LR reduction. As can be seen from figure~\ref{fig:projector}, this duration is sufficient for the weight to stabilize in noise-WD equilibrium. Also, 30GT corresponds to $\times3$ of Chinchilla \citep{hoffmann2022trainingcomputeoptimallargelanguage} compute optimal duration for 0.5B model scale, ensuring that the model is adequately trained. For the final validation, we use 240GT, a $\times24$ of compute-optimal duration to test the behavior of the multipliers in a more realistic setting of longer training duration. 

The other training hyperparameters also follow \citep{zuo2025falconh1familyhybridheadlanguage}. Specifically, we use a warmup duration of 0.1GT, batch size rampup with the square root LR scaling rule, and the global weight decay value of $0.1$, which is further modified by the tuned weight decay multipliers as given by table 8 of \citep{zuo2025falconh1familyhybridheadlanguage}. For each configuration of multipliers, we perform a learning rate sweep on a log-scale grid with $\sqrt{2}$ step and use the optimal value for experiments in the paper. 

\paragraph{Additional details for selected experiments.}
In all the experiments we use zloss \cite{chowdhery2022palmscalinglanguagemodeling} with coefficient $10^{-4}$ as it leads to better model performance. However, we disable the zloss for projector experiment discussed in section \ref{sec:whatlearned} and on figure \ref{fig:projector} as the zloss directly affects the behavior of the model logits, convoluting the clear interpretation of the logits norm required to investigate of learnable multipliers only. 

For width scaling experiment discussed in section \ref{sec:width_scaling}, we used a smaller model with 12 layers to access a wider range of widths within reasonable compute budget. The RMSNorms in all blocks were frozen to ensure unit normalization $\|\mathbf{x}\|=1$ of input block features, enabling cleaner interpretation of the observed norms.  

\section{MLP experiment}\label{sec:MLP_exp}
\begin{figure}
    \includegraphics[scale=0.41, clip, trim=10 10 10 10]{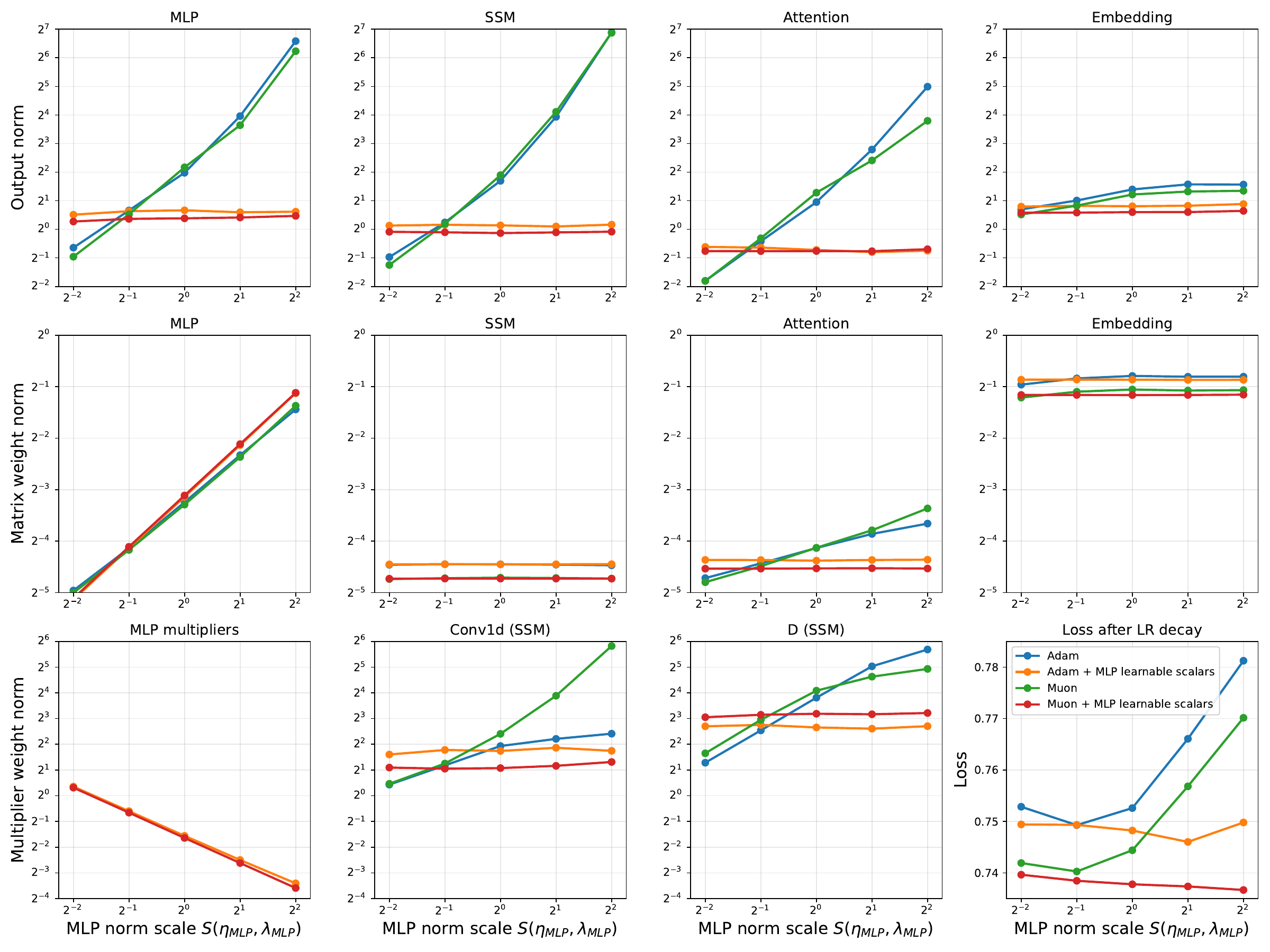}
    \caption[Short caption for list of figures]{%
All the subplots, except bottom right, show behavior of different norms as the MLP norm scale $S(\eta_{\mathrm{MLP}}, \lambda_{\mathrm{MLP}})$ is varied. The rows contain similar types of norms and share y-axis scale for easier comparison. The bottom right subplot duplicates the bottom right subplot of figure~\ref{fig:projector}; we add it for convenience as an illustration of performance across the four considered configurations.\\[0.5em]
\protect\textbf{(Top row):} The norms of the output of MLP, attention and SSM residual blocks, as well as the norm of the outputs of embedding layer. The norms are averaged across the tokens in a batch. Then, we apply geometric average to aggregate residual block norms across model layers.\\[0.5em]
\protect\textbf{(Middle row):} The norms of matrix weights: $W^\mathrm{up},W^\mathrm{gate},W^\mathrm{down}$ for MLP (see \eqref{eq:gated_mlp_fowardpass}), $W^{QKV}$\footnotemark and $W^{\mathrm{out}}$ for attention (see \eqref{eq:attention_forwardpass}), $W^{XZBCdt}$ and $W^{\mathrm{out}}$ for SSM (see \eqref{eq:SSM_forwardpass_1},\eqref{eq:SSM_forwardpass_2},\eqref{eq:SSM_forwardpass_3},\eqref{eq:SSM_forwardpass_4},\eqref{eq:SSM_forwardpass_5}), and the embeddings matrix. To aggregate the weight norms into a single metric for residual blocks, in addition to geometric averaging across the layers we further apply geometric averaging across the matrix types within the block, for example $W^\mathrm{up},W^\mathrm{gate},W^\mathrm{down}$ for MLP.\\[0.5em]
\protect\textbf{(Bottom row):} The norms of various non-matrix parameters that are free from the noise-WD equilibrium and can adjust their scale. For the three MLP multipliers we again apply geometric average; Conv1d acts as a row multiplier for $W^{XBC}$ in the SSM block (see \eqref{eq:SSM_forwardpass_1},\eqref{eq:SSM_forwardpass_2}); $D$ parameter scales the skip connection in the recurrent SSM computation, such that larger values of $D$ make SSM block computation closer to gated MLP.
}
    \label{fig:mlp}
\end{figure}

\footnotetext{For simplicity, we follow the implementation in our codebase and report the norm of the merged QKV matrix. Same applies to the merged $W^{XZBCdt}$ projection matrix of the SSM block}

In section~\ref{sec:whatlearned} and on figure~\ref{fig:projector} (bottom right) we observed that varying the scale MLP relative to all the other parts of the model leads to performance degradation while adding learnable multipliers to MLP removes this degradation. In this section, we illustrate that the degradation if indeed related to scale mismatch between mlp and other blocks  
In this section, we extend the discussion of MLP experiment presented in section~\ref{sec:whatlearned} and on figure~\ref{fig:projector} by showing the behavior of internal activations and parameter norms as we vary the norm scale $S(\eta_{MLP},\lambda_{MLP}) = \sqrt{\frac{\eta_{MLP}}{\lambda_{MLP}}}$ of the MLP matrix layers. 

\begin{figure}
    \centering
    \includegraphics[width=0.95\linewidth]{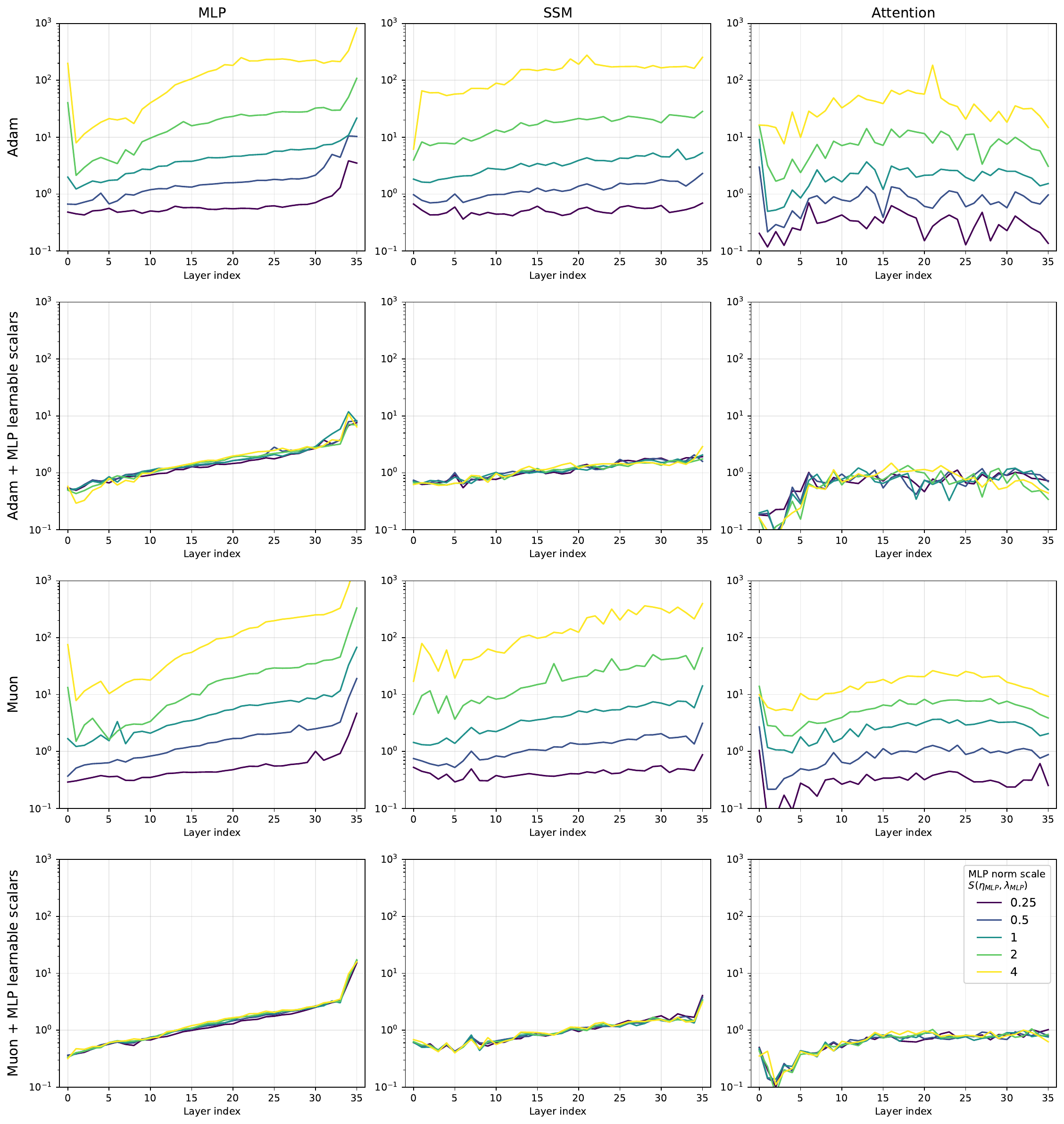}
    \caption{The behavior the norms of the MLP, attention and SSM residual blocks outputs across the model layers. This figure complements figure~\ref{fig:mlp}, where the norms were averaged across the layers. Overall, we observe strong layer-specific patterns, suggesting that the layer-averaged metrics in figure~\ref{fig:mlp} adequately capture the behavior of the norms with respect to MLP scale $S(\eta_{MLP},\lambda_{MLP})$.}
    \label{fig:mlp_layerwise}
\end{figure}

Let us first repeat and complement a brief description of the MLP experiment provided in section~\ref{sec:whatlearned}. First, we set RMSNorm weights $c_j\to1$ in \eqref{eq:rmsnorm} for all the MLP, attention, and SSM blocks to restrict the scale adaptation ability of these blocks. Then, we vary the norm scale of MLP block matrices in the following way: for each of $W^\mathrm{up},W^\mathrm{gate},W^\mathrm{down}$ (see \eqref{eq:gated_mlp_fowardpass}) we change learning rate $\eta$ and $\lambda$ while keeping $\eta\lambda=\mathrm{const}$. We denote $\eta_{MLP}=\eta/\eta_0$ and $\lambda_{MLP}=\lambda/\lambda_0$ the relative change of learning rate and weight decay, which are kept the same for all 3 MLP layers\footnote{The current absolute values of $\eta,\lambda$, as well as the baseline absolute values $\eta_0,\lambda_0$ might be different between $W^\mathrm{up}$, $W^\mathrm{gate}$, and $W^\mathrm{down}$ projections. This is the case for our experiment, where we use tuned $\mu P$ multipliers from \cite{zuo2025falconh1familyhybridheadlanguage} with slightly different multipliers for the three MLP projections.}. In figures~\ref{fig:projector},\ref{fig:mlp},\ref{fig:mlp_layerwise}, we use these relative values to focus on the change of the matrix norms\footnote{Same conventions apply for the projector experiment in figure~\ref{fig:projector}.}. The above fully describes our baseline configuration, and in the configuration with learnable MLP multipliers, we add three scalar multipliers $s^\mathrm{up},s^\mathrm{gate},s^\mathrm{down}$ to the three MLP matrix layers\footnote{Multiplicative symmetry makes one of the scalar multipliers $s^\mathrm{up}$ and $s^\mathrm{down}$ redundant. We keep both of them for simplicity, and, for the short training duration of MLP experiment, we did not observe any symmetry-induced training instabilities discussed in section~\ref{sec:symmetry}.}. These scalar multipliers had a fixed learning rate $\eta=10^{-2}$ and no weight decay. Finally, we run two versions of the experiment with Muon and Adam optimizers, while keeping all the other settings identical. 

With the experiment settings settled, we proceed to a discussion of the behavior of parameter and output norms as MLP matrix norm scaled $S(\eta_{MLP},\lambda_{MLP})$ is varied. First, we see a clear picture on figures \ref{fig:mlp} and \ref{fig:mlp_layerwise} for Adam and Muon \textbf{configurations with learnable multipliers}: 
\begin{enumerate}
    \item The norms of MLP matrices follow the equilibrium scale $S(\eta_{MLP},\lambda_{MLP})=\sqrt{\frac{\eta_{MLP}}{\lambda_{MLP}}}$.
    \item Yet, the outputs of all residual blocks stay constant regardless of the MLP matrices scale. 
    \item The constant level of residual blocks output is achieved thanks to MLP multiplier compensating the change in scale of MLP matrices (bottom left plot of figure~\ref{fig:mlp}). 
\end{enumerate}
This confirms the main thesis of the current work: the ability of learnable multipliers to freely adjust their scale in order to compensate a (presumably) suboptimal scale of its respective matrix layers, which is trapped in noise-WD equilibrium \eqref{eq:equilibrium_norm_scaling}.     

Next, we look at the \textbf{configurations without learnable multipliers} which turned out to be more nuanced. As we have restrained the scale adaptation ability of residual blocks by freezing the respective RMSNorm weights, pure equilibrium norm scaling \eqref{eq:equilibrium_norm_scaling} would create a imbalance between scales of MLP and attention/SSM blocks. This imbalance is expected to significantly degrade the model performance, which translates in a loss gap between configuration with and without learnable multipliers, seen on figure~\ref{fig:mlp} (bottom right). However, this imbalance also creates a significant optimization pressure to balance back the scales of MLP and attention/SSM blocks, and the model manages to align the scales of the residual blocks as can be seen in the top row of figure~\ref{fig:mlp}. This scale adaption is achieved via accumulating several mechanism which we list below. 
\begin{itemize}
    \item \textbf{MLP.} On fig.~\ref{fig:mlp} (middle left) we see that the norms of MLP matrix layers slightly deviate noise-WD scaling \eqref{eq:equilibrium_norm_scaling}. We interpret this deviation us a result of optimization pressure to reduce the scale gap between MLP and attention/SSM blocks. This introduces a strong enough gradient signal force that modifies the equilibrium \eqref{eq:equilibrium_norm_scaling} which was governed by gradient noise and WD forces only.
    \item \textbf{Attention.} We observe on fig.~\ref{fig:mlp} (top row, third column) the growth of the outputs of attention blocks with $S(\eta_{MLP},\lambda_{MLP})$. To estimate the norm of the attention block outputs, we may ignore the attention scores and approximate the norm as $\|W^{\mathrm{out}}W^V \mathbf{x}\|$ (see \eqref{eq:attention_forwardpass}). One way to increase this norm is to increase alignment between $W^{\mathrm{out}}$ and $W^V$, and also between $W^V$  and $\mathbf{x}$. Another way is for $W^{\mathrm{out}}$ or $W^V$ to escape noise-WD equilibrium, which is confirmed by the growth of the respective matrix norms on fig.~\ref{fig:mlp} (middle row, third column). We expect that value matrix is $W^V$ is more prone to escaping noise-WD equilibrium in group-query attention with large ratios of Q heads to KV heads: sharing of the KV heads results into more gradient signal coming through the matrix, resulting in higher signal-to-noise ratio. We suspect all these three mechanisms to increase the attention output to take place. Yet, from a slower growth of the attention outputs compared to SSM outputs we may conclude that employing these mechanisms negatively impacts the quality of attention blocks. 
    but slightly lags similar growth of SSM block output.
    \item \textbf{SSM.} Surprisingly, the outputs of SSM blocks grow at the same rate as outputs of MLP block while the norms of SSM matrices stay constant, in agreement with their noise-WD equilibrium values (fig.~\ref{fig:mlp} top and middle row, second column). After noticing this, we explored components of SSM block and identified two parts: \texttt{conv1d} (see \eqref{eq:SSM_forwardpass_1},\eqref{eq:SSM_forwardpass_2}) and SSM skip connection scale \texttt{D} that in fact play a role of learnable multipliers as they have vector-like shapes and hence not subject to noise-WD equilibrium. Indeed on fig.~\ref{fig:mlp} (bottom row, second and third columns) we observe the growth of these parameters, explaining the growth of SSM outputs at fixed norm of the respective matrix layers. 
    \item \textbf{Embedding.} The norm of embedding matrix stays constant, following its fixed equilibrium value (fig.~\ref{fig:mlp}, middle right), while the embedding outputs slightly grows to catch up with residual blocks outputs. To explain this growth, we note that that embedding output norm is a average norm of token embedding vectors weighted with frequency of each token in the data. Then, most frequent tokens display vector-like behavior with high signal-to-noise ratio in the gradients which allows them to partially escape noise-WD equilibrium and adjust their norm. On the other hand, low frequency tokens have low signal-to-noise ratio and therefore get trapped in the noise-WD equilibrium.
\end{itemize}

Finally, let us comment on the differences and similarities of Adam and Muon optimizers in the observed behaviors. In almost all the considered aspects, two optimizers behave identically, suggesting that noise-WD equilibrium mechanism could be a general phenomenon applicable to wide range of optimizer update rules. The cases where we observe a moderate difference between Adam and Muon include the growth of attention, \texttt{conv1d} and \texttt{D} norms. 

\paragraph{Remark on experiment design.} As we have seen above, the configuration without learnable multipliers manage to adapt the scale of attention and SSM blocks. This creates a complex picture of behaviors to ensure the growth of attention/SSM outputs, failing our original intent in restricting scale adaptation ability of these blocks to produce a clean separation between configurations with and without learnable multipliers. 

A few choices in the design of this MLP experiment may help to satisfy the original intention in the restricting the attention/SSM scale adaptability. For attention, we can switch from grouped query attention (GQA) to multi-head attention (MHA): this decreases the signal-to-noise ration in value matrices, making it harder for them to escape the equilibrium. For SSM, we can activate the internal RMSNorm\footnote{The reason we did not have this RMSNorm in our experiment is absence of it in Falcon-H1-0.5B architecture used throughout the work.}, as in \eqref{eq:SSM_forwardpass_4}, nullifying the effect of \texttt{conv1d} and \texttt{D} in the SSM output scale.        

\section{Multipliers placement}\label{sec:multipliers_placement}
In section \ref{sec:symmetry} we highlighted multiplicative and normalization symmetries that cause training instabilities if left unchecked. In this section, we write down the forward pass of the architecture employed in our experiments, Falcon-H1, in order to show such placement of learnable multipliers that removes all the redundancy, and associated symmetries, without reducing scale adaption expressivity of the multipliers. While in this section we focus on the specific hybrid attention-SSM architecture we used, we believe that it illustrates general principles guiding the symmetry-aware multiplier placement and can be easily generalized to other architectures.

However, when experimenting with such symmetry-aware placement, we observed a slightly worse performance than for configuration with all multipliers. Additionally, we note that there is no reliable way to fix residual normalization symmetry. Hence, in all our experiments, we chose to use both row and column multipliers while fixing the symmetry with multipliers weight decay, as discussed in section \ref{sec:symmetry}. In spite of using WD-based symmetry handling, we view symmetry-aware placement of multipliers as useful for better understanding of their effect, and potentially useful in future scenarios of LRMs usage.    

\subsection{Embedding and projector} For embedding, we use both column and vector multipliers. Theoretically, we could have removed all the multipliers from embedding to fix normalization residual symmetry. However, we observe that removing these multipliers results in a suboptimal performance. It could be hypothetically related to the dominant contribution of backbone blocks in the residual compared to the initial embeddings (see figure~\ref{fig:mlp}, top row), which results in block outputs ignoring the fixed embedding scale instead of adapting to it. 

For the projector, we again note that the well-established RMSNorm just before the projector already contains column multipliers for the projector matrix (see also projector experiment described in sec.~\ref{sec:whatlearned}). 

To summarize, we recommend adding both vector and column to the embedding layer while leaving the projector layer unchanged.

\subsection{Residual blocks}
We use a nowadays standard pre-LN design of the model blocks which uses RMSNorm in the beginning of each block. Specifically, for a residual $\mathbf{z}\in\mathbb{R}^d$ at the beginning of a block, the output $\mathbf{x}\in\mathbb{R}^d$\footnote{We slightly abuse the notation $\mathbf{x}$ (or, equivalently, $x_j$). In section~\ref{sec:learnable_multipliers} it denotes any internal activation just before application of a linear transformation by a matrix $\overline{W}$. In section~\ref{sec:whatlearned} it denotes final token features after applying RMSnorm normalization but before multiplying by the weights $c_j$. And, finally, in this section we include the RMSNorm weights $c_j$ inside $x_j$ to lighten the notations and to focus on the new learnable multipliers introduced to the classical architectures that already use RMSNorm layer.} of RMSNorm is given by
\begin{equation}\label{eq:rmsnorm}
    \Big(\operatorname{RMSNorm}(\mathbf{z})\Big)_j = \frac{c_j z_j}{\sqrt{\frac{1}{d}\sum_k z_k^2}}, 
\end{equation}
where $c_j$ are the learnable weights of the RMSNorm layer. As discussed in sections ~\ref{sec:learnable_multipliers} and \ref{sec:whatlearned}, $c_j$ act as column multipliers for the matrix layer that follows RMSNorm. In all the cases, we keep these ``column multipliers''.  

\paragraph{Gated MLP block.} The output of the block $y_i$ is computed as
\begin{equation}\label{eq:gated_mlp_fowardpass}
    y_i = \sum_k W^\mathrm{down}_{ik} \operatorname{SiLU}\Big(\sum_{j'} W^\mathrm{gate}_{kj'} x_{j'}\Big)\sum_j W^\mathrm{up}_{kj} x_j
\end{equation}
Then, a maximally expressive configuration of multipliers without symmetries would be 
\begin{itemize}
    \item Row and column multipliers for $W^\mathrm{down}$.
    \item No multipliers for $W^\mathrm{up}$.
    \item Row only multipliers for $W^\mathrm{gate}$.
\end{itemize}
Let us again remind that we don't include column multipliers for $W^\mathrm{gate}$ and $W^\mathrm{up}$ because of the previously assumed usage of RMSNorm weight. Note, however, that removing RMSNorm weight while adding a column multiplier for both $W^\mathrm{gate}$ and $W^\mathrm{up}$ would provide a non-redundant but more expressive configuration. At the moment, we have not tested this option (and similar options for the other blocks), leaving it for future work.  

\paragraph{Attention block.} The contribution $y_{i,l}^h$ of a single attention head $h$ to the block output at position $l$ reads
\begin{equation}\label{eq:attention_forwardpass}
    y_{i,l}^h = \sum_k W^{\mathrm{out},h}_{ik} \sum_{l'}\operatorname{Softmax}_{l'}\Big(\sum_{m} \sum_j W^{Q,h}_{mj} x_{j,l} \sum_{j'}W^{K,h}_{mj'}x_{j',l'}\Big) \sum_{j''} W^{V,h}_{kj''}x_{j'',l'}.
\end{equation}
Then, a maximally expressive configuration of multipliers without symmetries would be 
\begin{itemize}
    \item Row and column multipliers for $W^\mathrm{out}$.
    \item No multipliers for $W^V$ and $W^K$.
    \item Row only multipliers for $W^Q$.
\end{itemize}
Let us comment on this placement. Using row multipliers for both key and query matrices is redundant and was already discussed sec.~\ref{sec:symmetry}. Hence, we are left with the choice to put the multipliers either on the key or the query. A more expressive choice is dictated by the structure of the Group Query Attention (GQA) \citep{ainslie2023gqa}: a single key head is shared with several query heads. Therefore, putting multiplier os queries allows the model to learn per-head attention scales instead of per-group scales we would get in the case of attaching multipliers to keys. A similar reasoning applies to the choice of $W^\mathrm{out}$ column multipliers vs. $W^V$ row multipliers: attaching multipliers to output projection allows the model to learn with per-head output scales, while value projection multipliers would only learn per-group scale.           

\paragraph{SSM (Mamba2) block.} Taking into account a more complicated structure of the Mamba2 block, we break its computation into parts, focus on a single head, and, for simplicity, omit the temporal index and most of the internal channel indices. Then, Mamba2 forward pass reads

\begin{align}
\label{eq:SSM_forwardpass_1}
&\mathbf{X} = \operatorname{SiLU}\big(\operatorname{conv1d}\big(W^X \mathbf{x})\big)\big), \; \mathbf{Z} = \operatorname{SiLU}\big(W^Z \mathbf{x})\big), \\
\label{eq:SSM_forwardpass_2}
&\mathbf{B} = \operatorname{SiLU}\big(\operatorname{conv1d}\big(W^B \mathbf{x})\big)\big), \;
\mathbf{C} = \operatorname{SiLU}\big(\operatorname{conv1d}\big(W^C \mathbf{x})\big)\big), \\
\label{eq:SSM_forwardpass_3}
&\mathbf{dt} = \operatorname{Softplus}\big(W^{dt} \mathbf{x}+\mathbf{b}_dt\big), \\    
\label{eq:SSM_forwardpass_4}
& \mathbf{F} = \operatorname{RMSNorm}\Big(\operatorname{SSM}\big(\mathbf{X}, \mathbf{B}, \mathbf{C}, \mathbf{dt}\big) \odot \mathbf{Z}\Big),\\
\label{eq:SSM_forwardpass_5}
& y_i = \sum_j W^{\mathrm{out}}_{ik} F_k.
\end{align}
Here $\operatorname{SSM}\big(\mathbf{X}, \mathbf{B}, \mathbf{C}, \mathbf{dt}\big)$ is the mamba2 sequence transformation \citep{dao2024transformersssmsgeneralizedmodels}, and $\operatorname{conv1d}(\cdot)$ is casual per-channel convolution. Let us comment on this structure to arrive at our final multiplier configuration. RMSNorm layer is typically used, but can be skipped in some cases, including \texttt{Falcon-H1-0.5B} architecture that we use for our experiments.  

Let us comment on each part of the computation to determine where learnable multipliers are required. Casual conv1d can be viewed as a more expressive operation than row multipliers, as it adds short-range temporal mixing in addition to the per-channel rescaling. Therefore, we do not apply row multipliers to $W^X, W^B, W^C$. Since the output of both $W^{dt}$ and $W^Z$ goes into a non-linearity, and the respective parts do not have native parameters able to learn the scale, we apply row multipliers to these matrices. Finally, output projection $W^{\mathrm{out}}$ does not have any symmetries associated with it, and thus requires both row and column multipliers unless the RMSNorm layer is present and already contains the column multiplier. Summarizing, we have  
\begin{itemize}
    \item Row only multiplier for $W^{\mathrm{out}}$. Column multiplier is added if the internal RMSNorm layer is skipped.
    \item No multipliers for $W^X,W^B,W^C$.
    \item Row only multiplier for $W^Z$ and $W^{dt}$.
\end{itemize}

\section{Additional plots}\label{sec:additionalplots}

\begin{figure}[H]               
  \centering
    \includegraphics[scale=0.64]{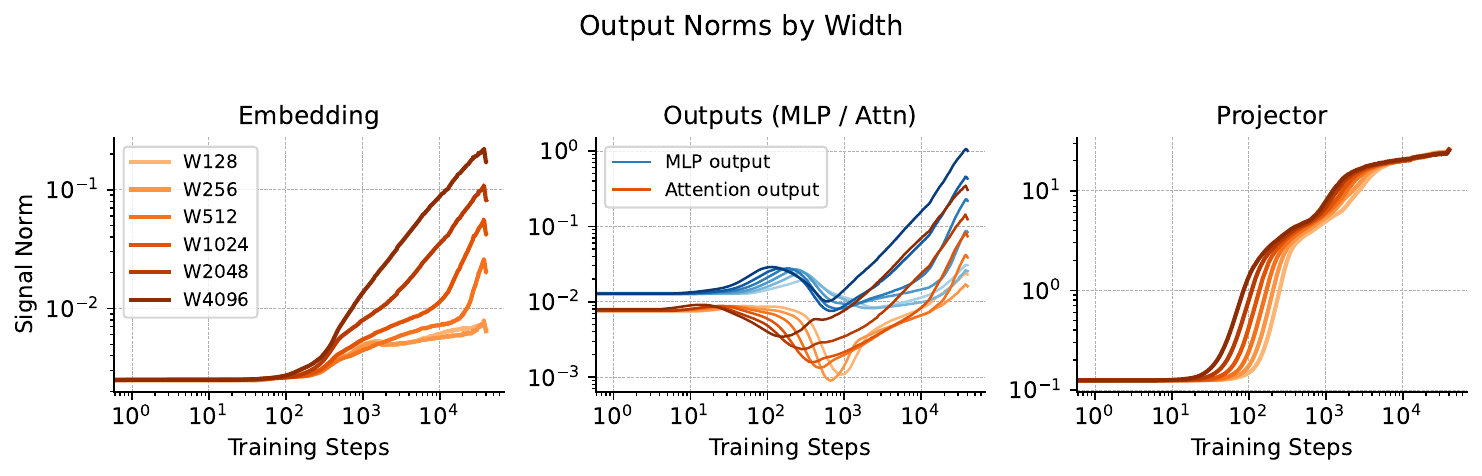}
  \caption{Evolution of signal norms during training across different model widths ($W128$ to $W4096$). 
  \textbf{(Center)} MLP and Attention output norms show consistent norm growth in the later stages of training, which we attribute to drift along the direction of residual normalization symmetry. 
  \textbf{(Left)} Embedding output norms also grow to match residual growth. 
  \textbf{(Right)} Projector output norms also grow with training time but seem to plateau at the same level, presumably corresponding to the reasonable logits scale. The model width affect the projector output norm only in the intermediate stage of training, while for MLP and attention outputs norm are grow with different offset for different widths, reflecting arbitrary scale of residuals due to normalization symmetry.}
  \label{fig:width_mup_out_norms}
\end{figure}

\begin{figure}[h]               
  \centering
    \includegraphics[scale=0.48]{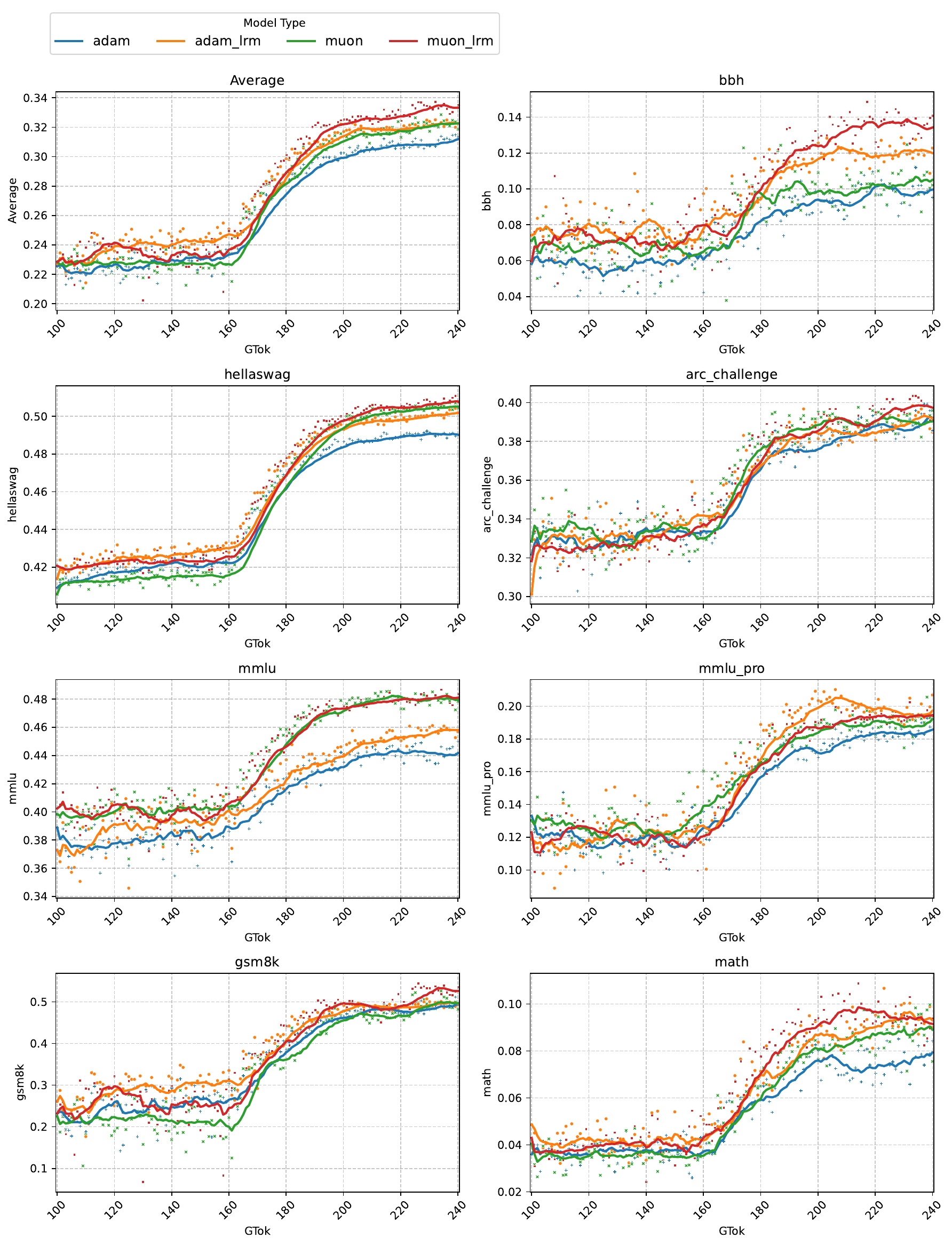}
  \caption{The detailed evaluation curves for the long runs reported in table~\ref{tab:evals}. The markers correspond to actual evaluation score at a given checkpoint while solid lines denote running window average score over 20 last checkpoints. We evaluated checkpoints every gigatoken to carefully average the benchmarks stochasticity, and started this frequent evaluation only from 100GT due to compute constraints. We perform $\times 32$ exponential decay from 160GT to 200GT, which explains the growth of the scores in this time window is thanks to the learning rate decay. After the end of exponential decay, the model was trained for 40 more gigatokens with minimal learning rate to obtain enough evaluation points ensuring well averaged scores reported in table~\ref{tab:evals}.}
  \label{fig:eval_trajectories}
\end{figure}

\end{document}